\documentclass[oneside,11pt]{article}

\usepackage{url,mathrsfs,algorithm}
\usepackage{amsmath,wrapfig,color}
\usepackage{amsfonts}
\usepackage{graphicx}
\usepackage{subfigure}

\begin{document}
\title{\Huge Generalized Intersection Kernel}

\author{ \bf{Ping Li} \\
         Department of Statistics and Biostatistics\\
         Department of Computer Science\\
       Rutgers University\\
          Piscataway, NJ 08854, USA\\
       \texttt{pingli@stat.rutgers.edu}\\
}

\date{}

\maketitle

\begin{abstract}
\noindent Following the very recent line of work on the ``generalized min-max'' (GMM) kernel~\cite{Report:Li_GMM16}, this study  proposes the ``generalized intersection'' (GInt) kernel and the  related ``normalized generalized min-max'' (NGMM) kernel. In computer vision, the (histogram) intersection kernel has been popular, and the  GInt kernel generalizes it to data which can have both negative and positive entries. Through an extensive empirical classification study on 40 datasets from the UCI repository, we are able to show that this (tuning-free) GInt kernel performs fairly well. \\

\noindent The empirical results also demonstrate that the NGMM kernel typically outperforms the GInt kernel. Interestingly, the NGMM kernel has another interpretation ---  it is the ``asymmetrically transformed'' version of the GInt kernel, based on the  idea of ``asymmetric hashing''~\cite{Proc:ALSH_WWW15}. Just like the GMM kernel, the NGMM kernel can be efficiently linearized through (e.g.,) generalized consistent weighted sampling (GCWS), as  empirically validated in our study. Owing to the discrete nature of  hashed values, it also provides a scheme for approximate near neighbor search.\\

\noindent The proposed GInt kernel and NGMM kernel provide additional options for practitioners to deal with  data in their specific domains. We also note that there  have been a series of (unpublished) works on related topics. For example, \cite{Report:Li_GMM16} proposed the GMM kernel and the ``normalized random Fourier features (NRFF)'' method; \cite{Report:Li_GMM_Nys16} compared the Nystrom method for approximating the GMM kernel with NRFF; \cite{Report:Li_GMM_Theory16} developed some basic mathematical theory (e.g., the convergence and asymptotic normality) for the GMM kernel;  ~\cite{Report:Li_SignStable15} compared the hashed (linearized) GMM kernel with   a series of kernels which can be approximated by sign stable random projections.

\end{abstract}

\section{Introduction}

Following the idea from the recently proposed ``generalized min-max (GMM)'' kernel~\cite{Report:Li_GMM16}, we propose two types of nonlinear kernels which are basically tuning free and can handle data vectors with both negative and positive entries. The first step is a simple transformation on the original data. Consider, for example,   the original data vector $u_i$, $i=1$ to $D$. We define the following transformation, depending on whether an entry $u_i$ is positive or negative:
\begin{align}\label{eqn_transform}
 \left\{\begin{array}{cc}
\tilde{u}_{2i-1} = u_i,\hspace{0.1in} \tilde{u}_{2i} = 0&\text{if } \ u_i >0\\
\tilde{u}_{2i-1} = 0,\hspace{0.1in} \tilde{u}_{2i} =  -u_i &\text{if } \ u_i \leq 0
\end{array}\right.
\end{align}
For example, when $D=2$ and $u = [-5\ \ 3]$, the transformed data vector becomes $\tilde{u} = [0\ \ 5\ \ 3\ \ 0]$.

Once we have only nonnegative data, we can define the GMM kernel as proposed in~\cite{Report:Li_GMM16}
\begin{align}
GMM(u,v) = \frac{\sum_{i=1}^{2D}\min\{\tilde{u}_i,\tilde{v}_i\}}{\sum_{i=1}^{2D} \max\{\tilde{u}_i,\tilde{v}_i\}}
\end{align}

Inspired by the above idea, a variety of nonlinear kernels can be analogously defined, for example, the ``generalized intersection (GInt)'' kernel:
\begin{align}
GInt(u,v) = \sum_{i=1}^{2D} \min\{\tilde{u}_i,\tilde{v}_i\},\hspace{0.5in} \sum_{i=1}^{2D} \tilde{u}_i = \sum_{i=1}^{2D}\tilde{v}_i = 1
\end{align}
and the ``normalized GMM (NGMM)'' kernel:
\begin{align}
NGMM(u,v) = \frac{\sum_{i=1}^{2D}\min\{\tilde{u}_i,\tilde{v}_i\}}{\sum_{i=1}^{2D} \max\{\tilde{u}_i,\tilde{v}_i\}},\hspace{0.5in} \sum_{i=1}^{2D} \tilde{u}_i = \sum_{i=1}^{2D}\tilde{v}_i = 1
\end{align}

Note that original (histogram) intersection kernel has been a popular tool in computer vision~\cite{Maji_CVPR08}. In this study, we will provide an extensive empirical evaluation of the GInt kernel and NGMM kernel on 40 classification datasets from the UCI repository. The empirical results indicate that the NGMM kernel typically outperforms the GInt kernel. In addition, there is an interesting connection between these two kernels, because we can re-write the NGMM kernel as
\begin{align}
NGMM(u,v) = \frac{\sum_{i=1}^{2D}\min\{\tilde{u}_i,\tilde{v}_i\}}{2-\sum_{i=1}^{2D} \min\{\tilde{u}_i,\tilde{v}_i\}},\hspace{0.5in} \sum_{i=1}^{2D} \tilde{u}_i = \sum_{i=1}^{2D}\tilde{v}_i = 1
\end{align}
which means that the NGMM kernel is a monotonic transformation of the GInt kernel. This connection can be also be interpreted as that the NGMM kernel is an ``asymmetrically transformed'' version of the GInt kernel, based on the recent idea of asymmetric hashing~\cite{Proc:ALSH_WWW15}.\\

Next, we first provide an empirical study on kernel SVMs based on the aforementioned kernels, followed by an empirical study on hashing (linearizing) the NGMM kernel.

\section{An Experimental Study on Kernel SVMs}\label{sec_kernel}

\begin{figure}[h!]
\begin{center}
\mbox{
\includegraphics[width=2.4in]{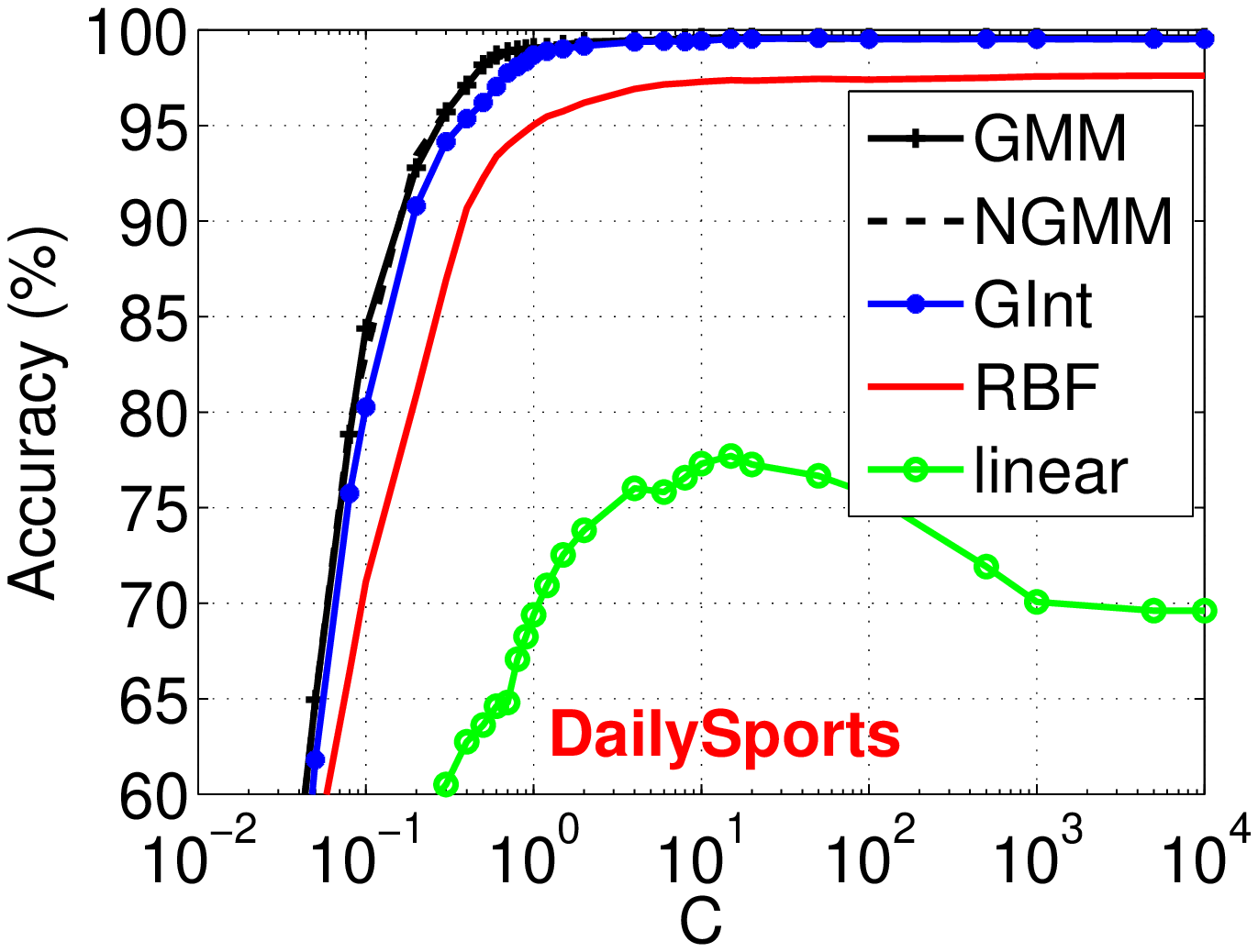}\hspace{-0.2in}
\includegraphics[width=2.4in]{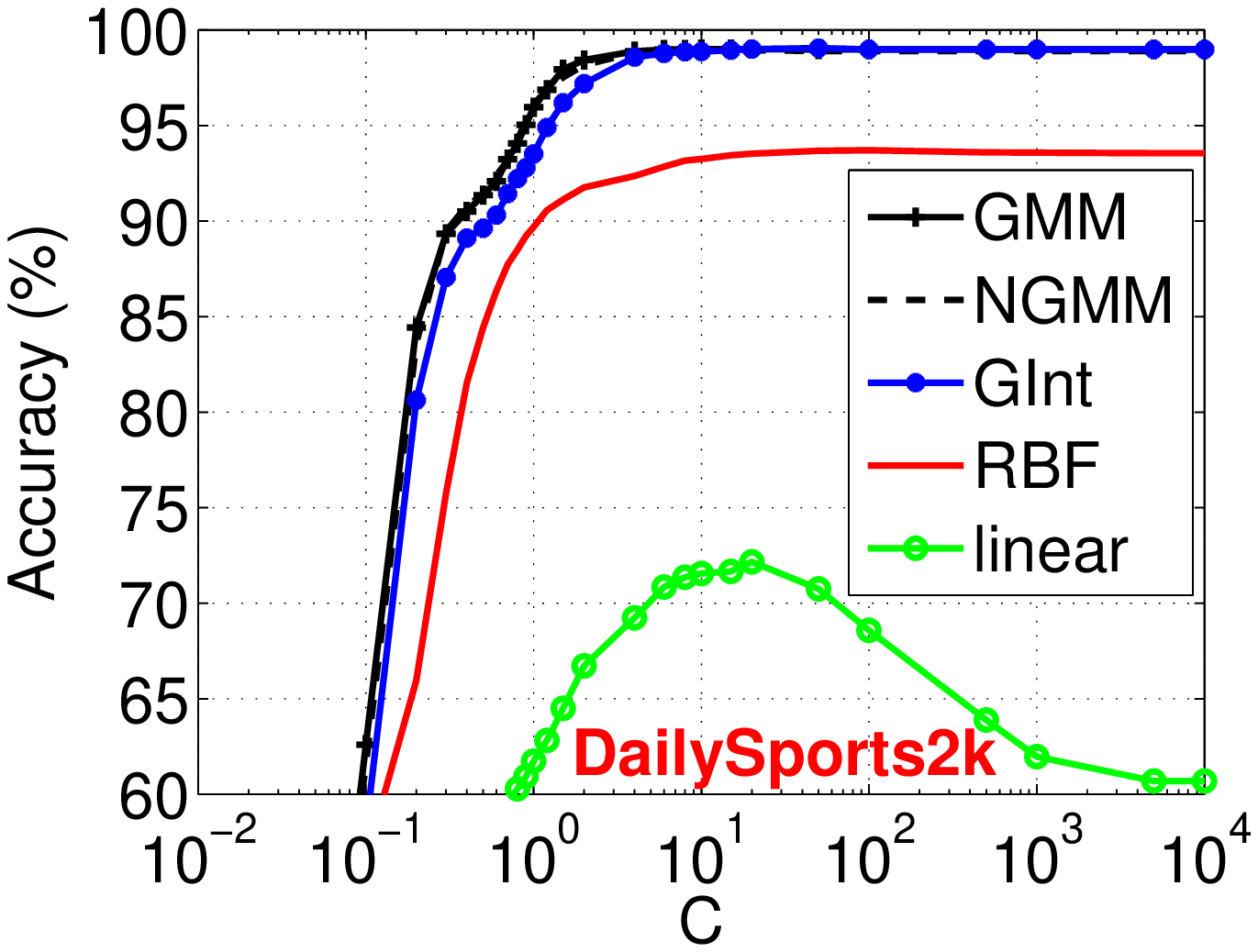}\hspace{-0.2in}
\includegraphics[width=2.4in]{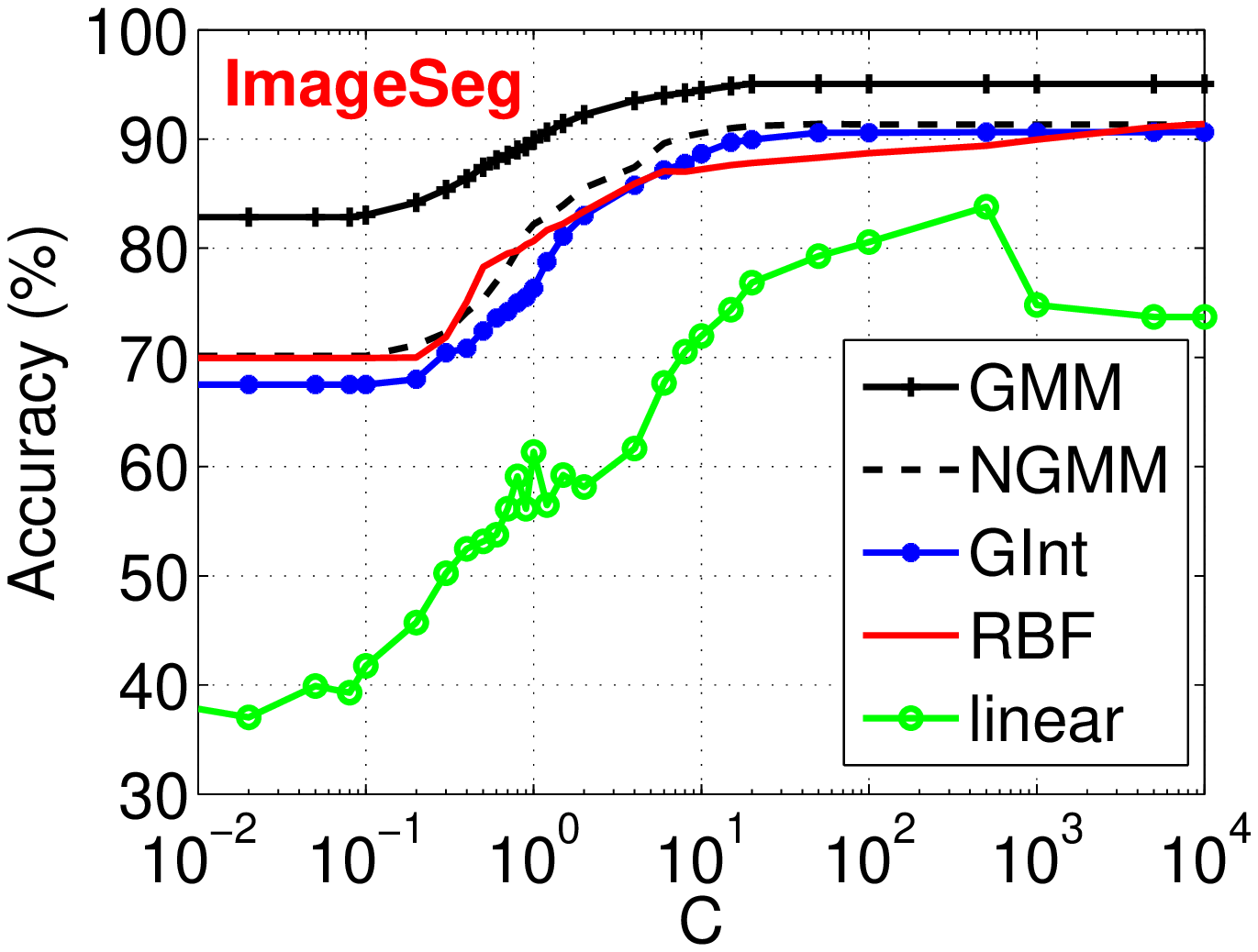}
}

\mbox{

\includegraphics[width=2.4in]{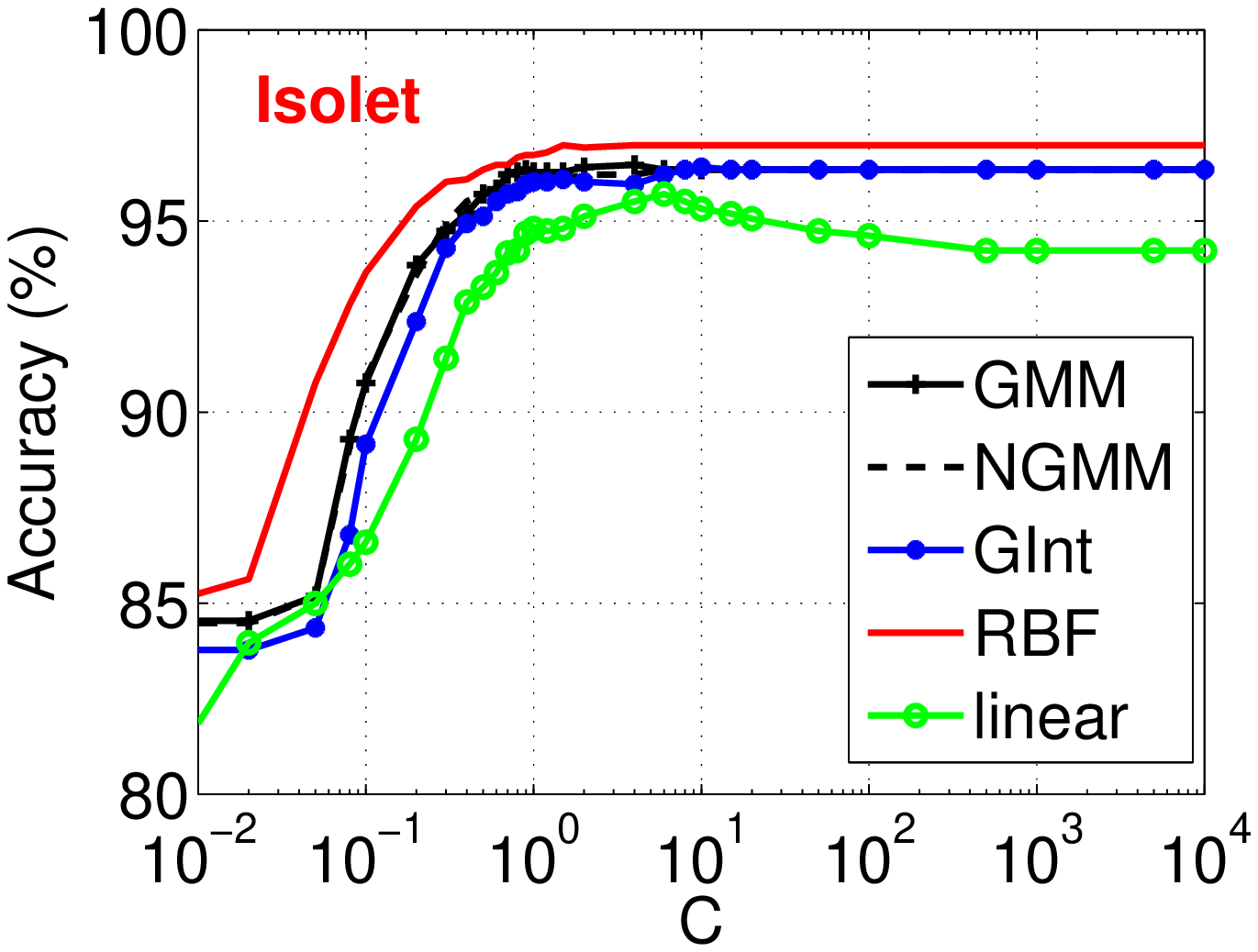}\hspace{-0.2in}
\includegraphics[width=2.4in]{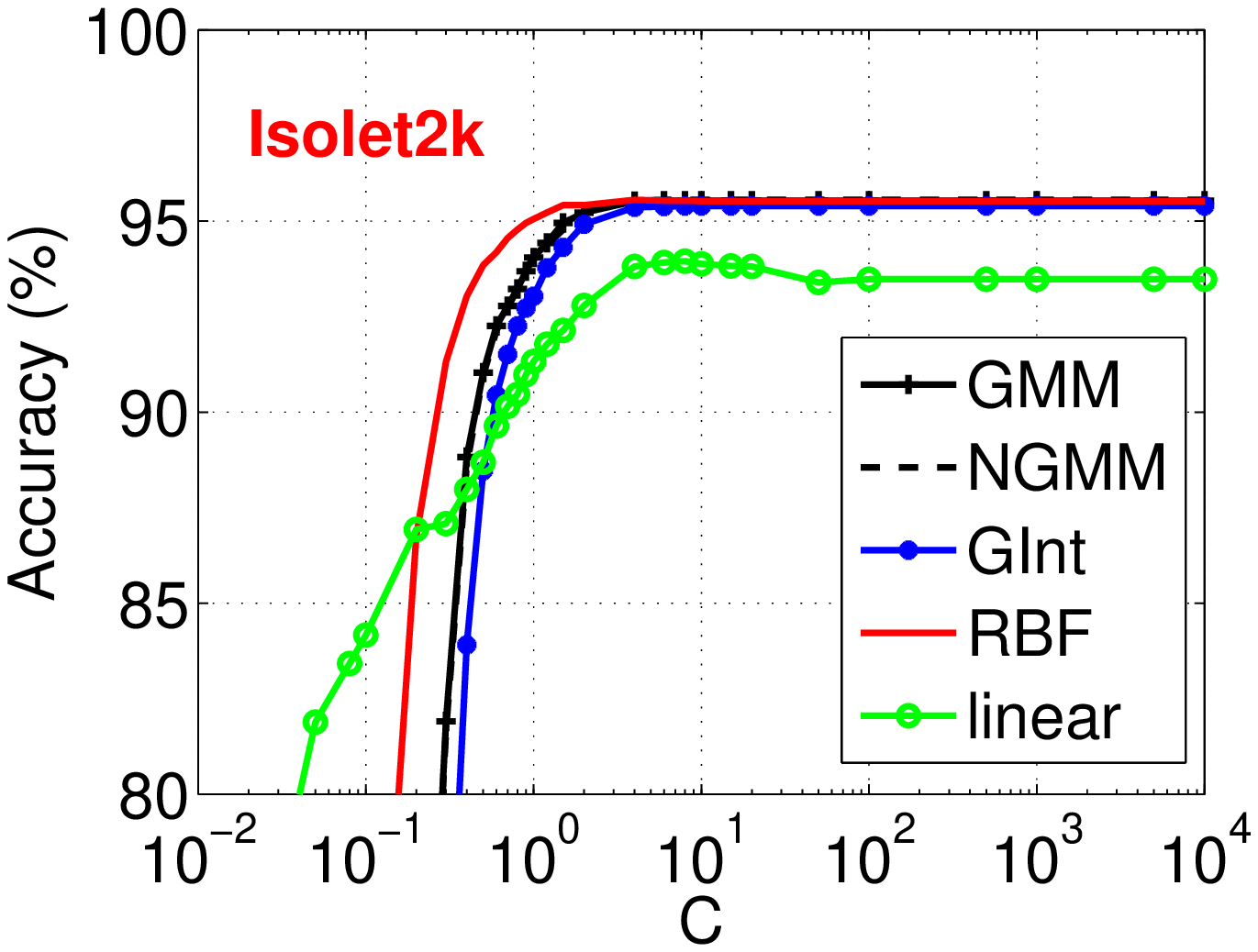}\hspace{-0.2in}
\includegraphics[width=2.4in]{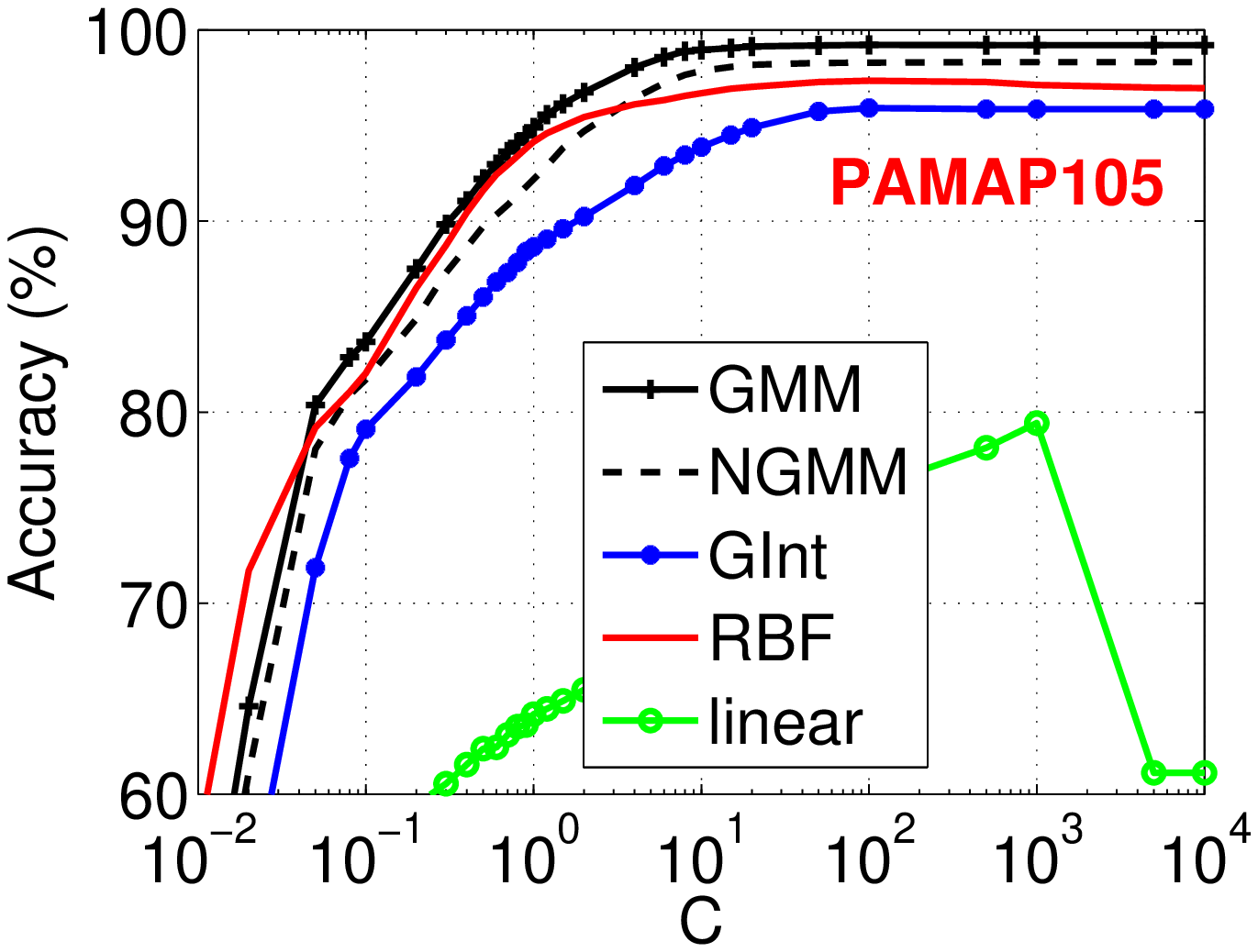}
}

\mbox{

\includegraphics[width=2.4in]{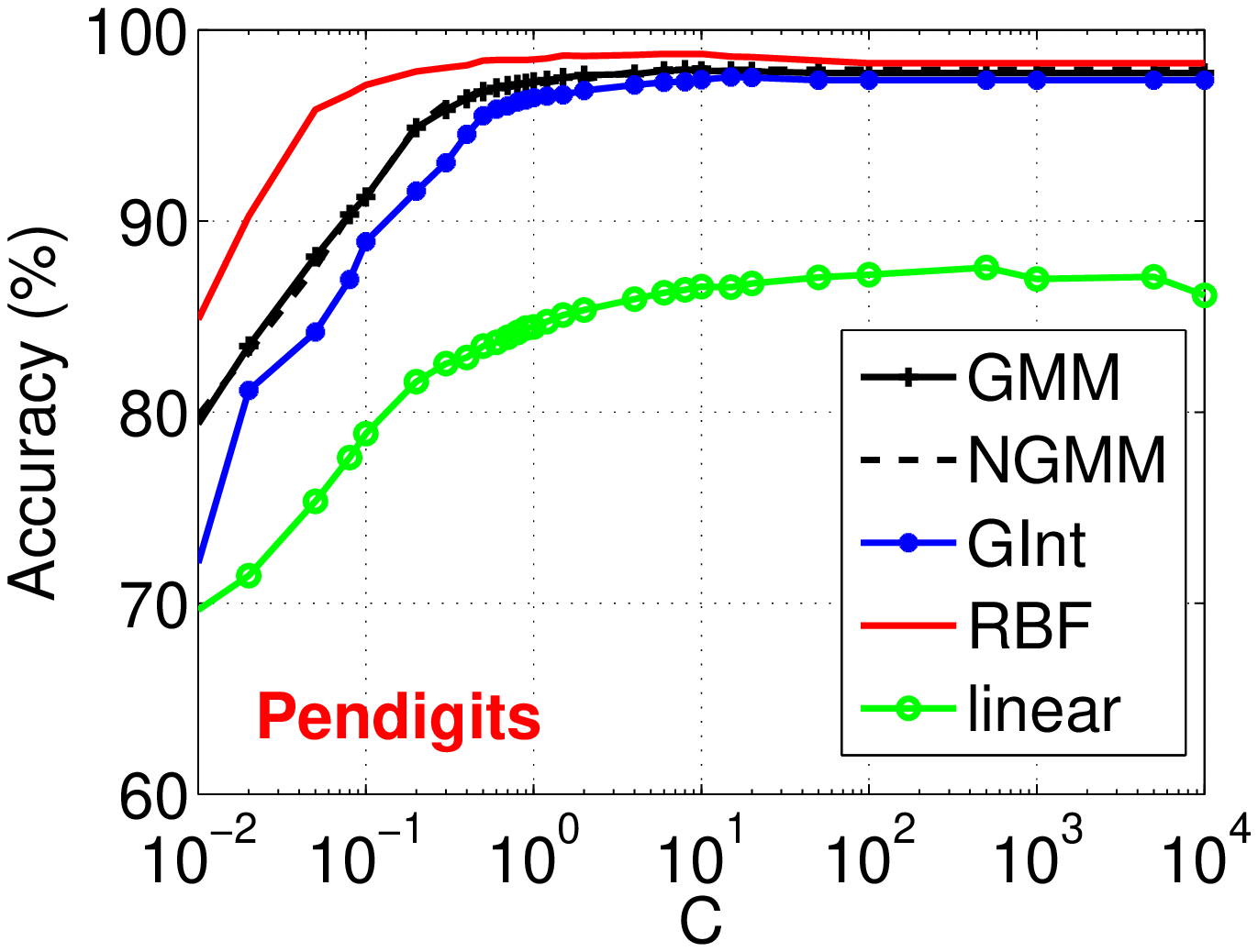}\hspace{-0.2in}
\includegraphics[width=2.4in]{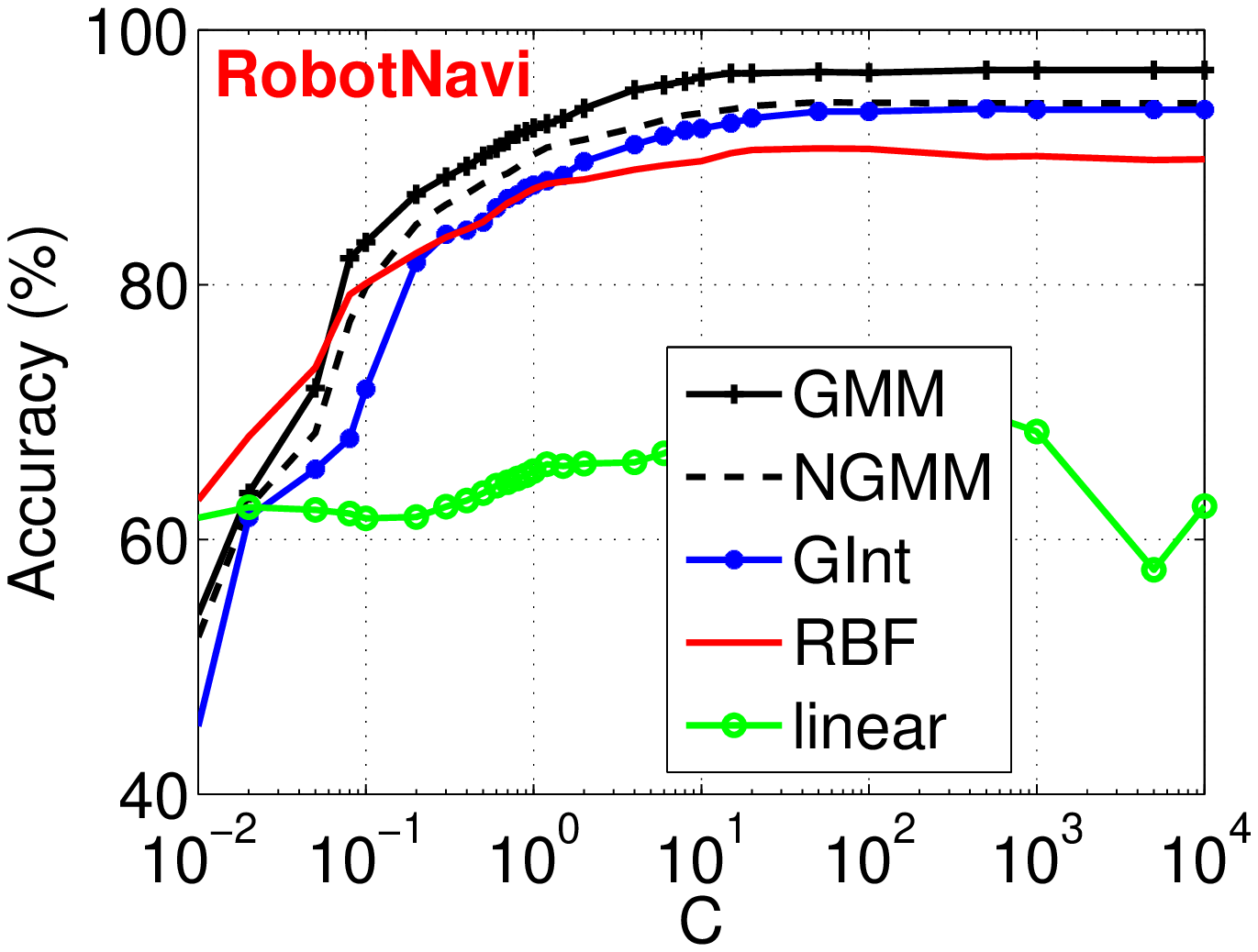}\hspace{-0.2in}
\includegraphics[width=2.4in]{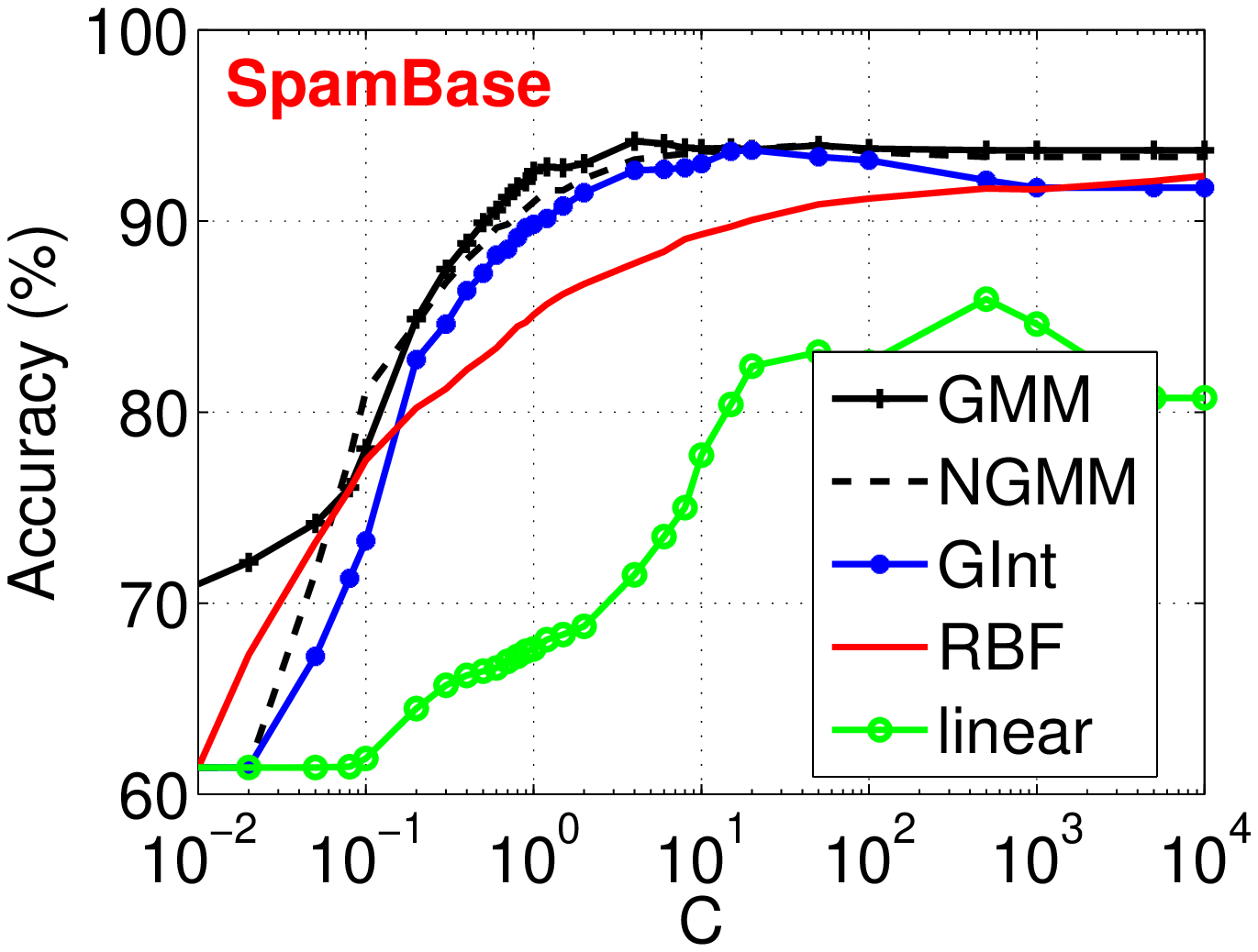}
}
\mbox{
\includegraphics[width=2.4in]{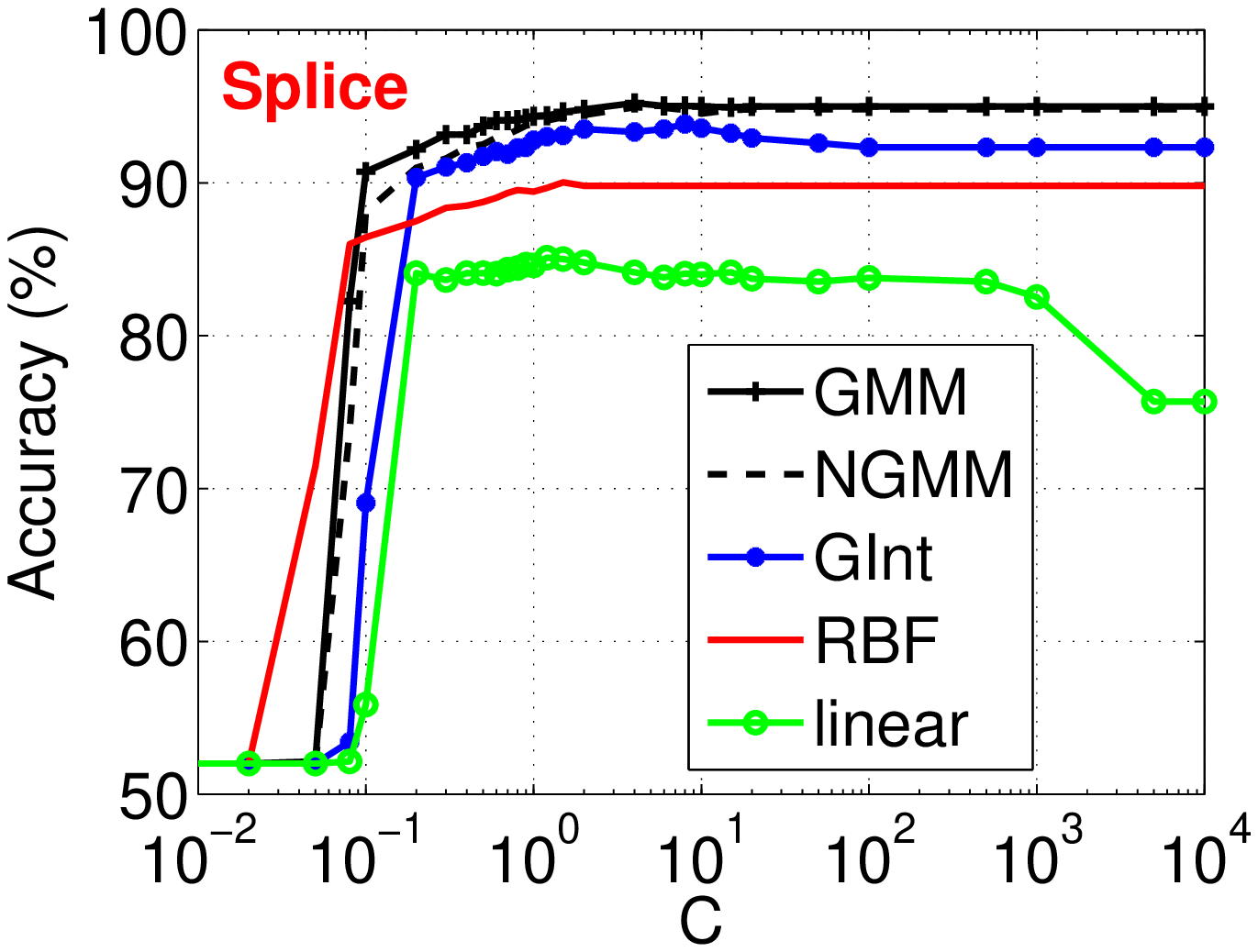}\hspace{-0.2in}
\includegraphics[width=2.4in]{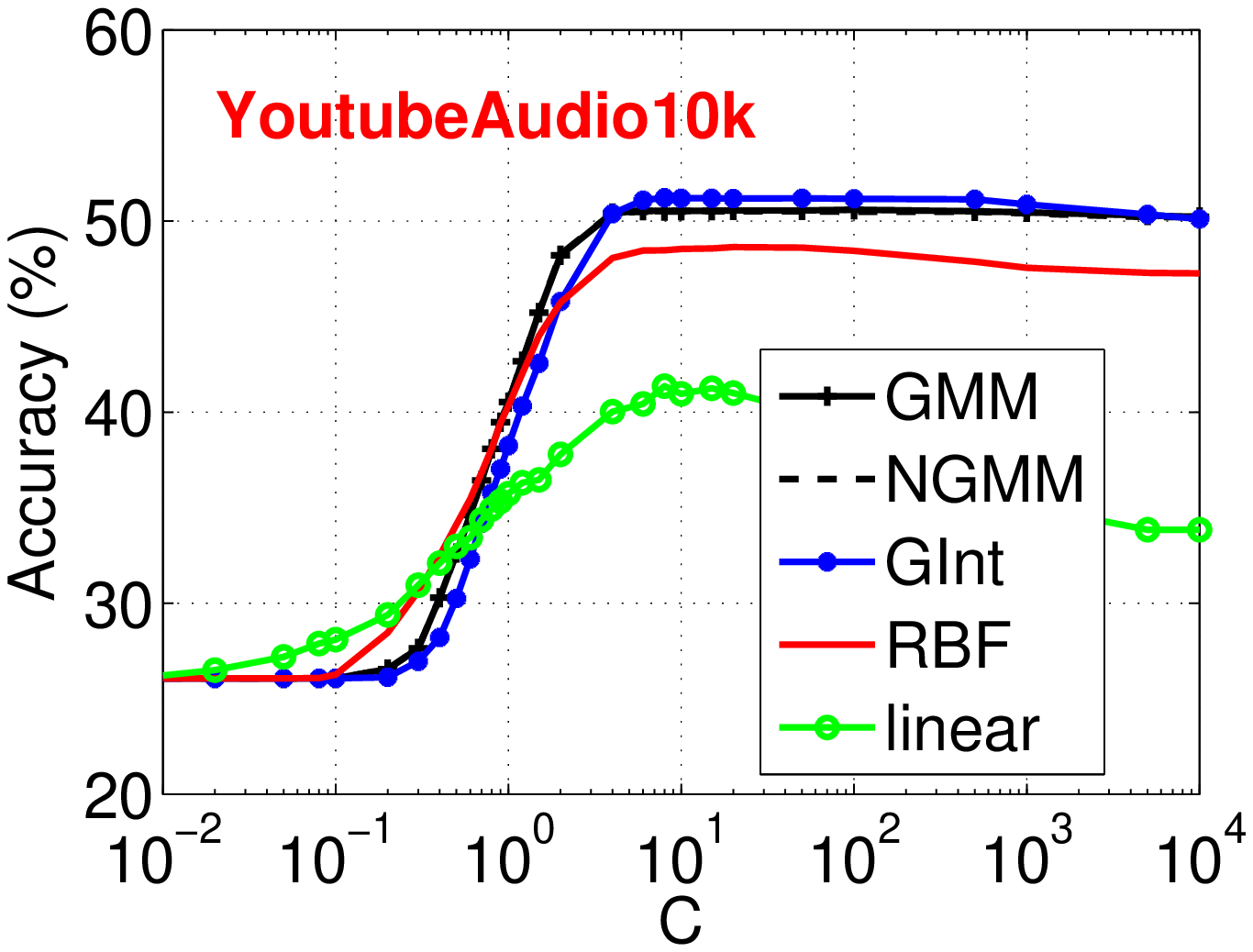}\hspace{-0.2in}
\includegraphics[width=2.4in]{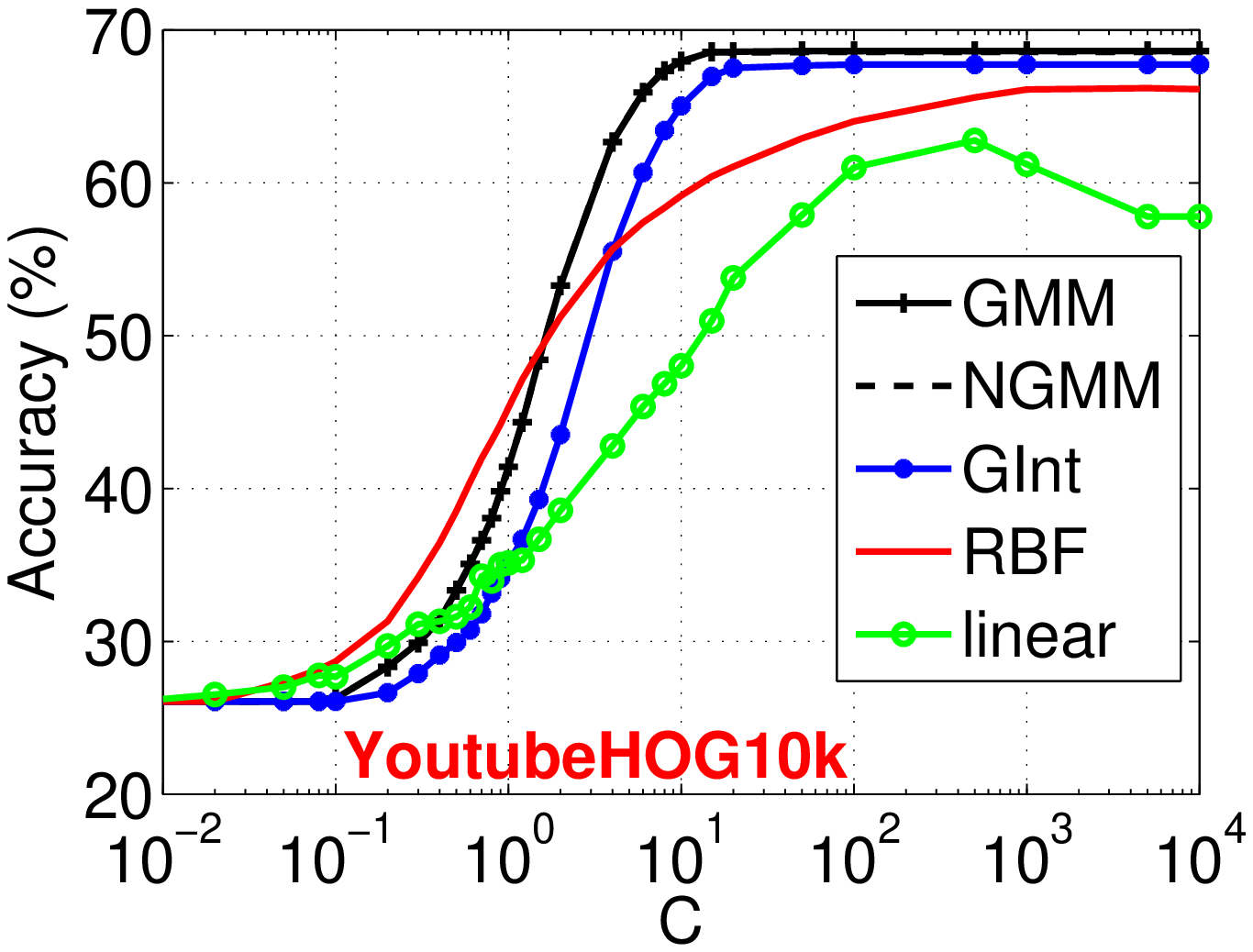}
}

\end{center}
\vspace{-0in}
\caption{Test classification accuracies of various kernels using LIBSVM pre-computed kernel functionality. The results are presented  with respect to $C$, which is the $l_2$-regularized kernel SVM parameter. Note that except for the RBF kernels, all kernels are tuning-free. For RBF, at each $C$, we report the best test accuracies from a wide range of kernel parameter ($\gamma$) values.}\label{fig_GInt}

\end{figure}

Table~\ref{tab_data} lists 40 publicly available  datasets, solely from the UCI repository, for our experimental study, along with the kernel SVM classification results for the RBF kernel, the  GMM kernel, and the proposed NGMM kernel and GInt kernel, at the (individually) best $l_2$-regularization $C$ values. More detailed results (for all regularization $C$ values) are available in Figure~\ref{fig_GInt}. To ensure repeatability, we use the LIBSVM pre-computed kernel functionality. Note that for the RBF kernel with a  scale parameter $\gamma$, we report  the best result among a wide range of  the $\gamma$  and $C$ values. \\

The classification results in Table~\ref{tab_data} and Figure~\ref{fig_GInt} indicate that, on these datasets, the GInt kernel significantly outperforms the linear kernel, confirming the advantage of exploring data nonlinearity. The NGMM kernel typically outperforms the GInt kernel. It appears that the GInt kernel can be fairly safely replaced by the NGMM kernel, especially as the NGMM kernel can efficiently computed.

\begin{table}[h!]
\caption{\textbf{40 public (UCI) datasets  and kernel SVM results}. We report the test classification accuracies for the linear kernel, the best-tuned RBF kernel, the GMM kernel, the NGMM kernel, and the GInt kernel, at the best SVM regularization  $C$ values.
}
\begin{center}{
{\begin{tabular}{l r r r c l c c c}
\hline \hline
Dataset     &\# train  &\# test  &\# dim &linear  &RBF  &GMM &NGMM &GInt\\
\hline
DailySports &4560 &4560 &5625  & 29.47 &97.61 &\textbf{99.61} &99.54 &99.56\\
DailySports2k&2000&7120&5625& 72.16  &93.71   &98.99   &98.93  &{\bf99.05}\\
Gesture &4937 &4936 &32 & 37.22   &61.06   &{\bf65.50}  &58.93 &47.02\\
ImageSeg &210 &2100 &19 &83.81   &91.38   &{\bf95.05}   &91.38  & 90.62\\
Isolet &6238 &1559 &617 &95.70 &{\bf96.99} & 96.47 &96.34 &96.41\\
Isolet2k &2000 &5797 &617 & 93.95   &{\bf95.55} &95.53 &{\bf95.55}&95.39\\
MSD20k &20000 &20000 &90 &66.7 &68.07 &\textbf{71.1} &68.30 &67.57 \\
MHealth20k&20000&20000&23&72.62   &82.38   &{\bf85.28}   &84.18   &81.89\\
MiniBooNE20k&20000&20000&50&88.42   &92.83   &{\bf93.00}   &92.79  &92.20\\
Magic &9150 &9150 &10 &78.04   &83.90  &{\bf87.02}   &84.10  & 81.75\\
Musk &3299 &3299 &166  & 95.09 &\textbf{99.33} &99.24 &99.12 &99.09\\
Musk2k&2000&4598&166&94.80   &97.63   &{\bf98.02}   &{\bf98.02}   &97.85\\
Optdigits &3823 &1797 &64 & 95.27   &{\bf98.72}   &97.72   &97.44  &96.77\\
PageBlocks &2737  &2726 &10 &95.87   &{\bf97.08}   &96.56   &{\bf97.08} &97.04\\
Parkinson &520&520&26&61.15   &66.73   &{\bf69.81}   &62.12  &63.65\\
PAMAP101 &20000 &20000 &51 &76.86   &96.68  &{\bf98.91}   &97.24      &93.07\\
PAMAP102 &20000 &20000 &51 &81.22   &95.67  &{\bf98.78}  &96.56   &93.64\\
PAMAP103 &20000 &20000 &51 & 85.54  &97.89  &{\bf99.69}  &98.75  & 96.62\\
PAMAP104 &20000 &20000 &51 &84.03  &97.32   &{\bf99.30}  &97.95  &95.82\\
PAMAP105 &20000 &20000 &51 &79.43  &97.34   &{\bf99.22}  &98.31  &95.91\\
Pendigits &7494 &3498 &16 &87.56   &{\bf98.74}   &97.91   &98.00  &97.54\\
RobotNavi &2728 &2728 &24 &69.83   &90.69   &{\bf96.85}   &94.32  &93.81\\
Satimage &4435 &2000 &36 &72.45   &85.20   &{\bf90.40}   &83.50   &83.15\\
SEMG1 &900 &900 &3000  &26.00   &\textbf{43.56}   &41.00 &41.22 &38.33\\
SEMG2 &1800 &1800 &2500  &19.28  &29.00    &\textbf{54.00} &51.00 &51.06\\
Sensorless &29255 &29254 &48 &61.53 &90.83  &{\bf99.39}   &92.62   &92.79\\
Shuttle500 &500 &14500 &9 &91.81   &99.52  &{\bf99.65}   &99.61  &99.59\\
SkinSeg10k&10000&10000&3& 93.36   &99.74 &{\bf99.81}   &99.74   &99.74\\
SpamBase &2301&2300&57&   85.91&   92.38 & {\bf94.17}   &94.00 & 93.70\\
Splice &1000&2175&60&85.10   &90.02   &{\bf95.22}   &94.94    &93.84\\
Theorem &3059&3059&51&   67.83   &70.48   &{\bf71.53}   &70.97   &70.06\\
Thyroid &3772 &3428 &21 &95.48   &97.20   &{\bf98.31}   &97.14   &97.17\\
Thyroid2k &2000&5200&21&   94.90   &96.98  &{\bf98.40}   &97.25   &97.00\\
Urban &168&507&147&   62.52   &50.30  &{\bf66.08}   &57.60 &58.98\\
Vertebral &155&155&6&   80.65   &83.23 &{\bf89.02}   &81.29&81.94\\
Vowel &264 &264 &10 &39.39 &94.70  &{\bf96.97}  &89.39  &85.61\\
Wholesale &220&220&6&   89.55   &90.91   &{\bf93.18}   &89.55   &89.55\\
Wilt &4339&500&5&62.60   &83.20   &{\bf87.20}  &82.60 & 81.00\\
YoutubeAudio10k&10000&11930&2000&   41.35   &48.63   &50.59   &50.51  &{\bf51.22}\\
YoutubeHOG10k&10000&11930&647&   62.77   &66.20   &{\bf68.63}   &68.56 &67.73\\
YoutubeMotion10k&10000&11930&64& 26.24  &28.81   &{\bf31.95}   &31.94   &30.14\\
YoutubeSaiBoxes10k&10000&11930&7168&46.97   &49.31   &51.28   &51.17 &{\bf51.48}\\
YoutubeSpectrum10k&10000&11930&1024&26.81   &33.54   &{39.23}   &{\bf39.27} &35.98\\
\hline\hline
\end{tabular}}
}
\end{center}\label{tab_data}

\end{table}
\newpage\clearpage

\newpage\clearpage

\section{Kernel Linearization}

The kernel classification experiments  in Table~\ref{tab_data} and Figure~\ref{fig_GInt}  have demonstrated the effectiveness of nonlinear kernels (GInt, NGMM, GMM, and RBF) in terms of prediction accuracies, compared to the linear kernel.  However, it is well understood~\cite{Book:Bottou_07} that computing kernels are expensive and the kernel matrix, if fully materialized, does not fit in memory even for relatively small applications.

For example, for  a small dataset with merely $60,000$ data points, the $60,000 \times 60,000$ kernel matrix  has $3.6\times 10^9$ entries.  In practice, being able to linearize nonlinear kernels becomes highly beneficial. Randomization (hashing) is a popular tool for kernel linearization. After data linearization, we can then apply our favorite linear learning packages such as LIBLINEAR~\cite{Article:Fan_JMLR08} or SGD (stochastic gradient descent)~\cite{URL:Bottou_SGD}.

There are multiple ways for kernel linearization. See~\cite{Report:Li_GMM_Nys16} for the work on utilizing  the  Nystrom method for kernel approximation~\cite{Article:Nystrom1930,Proc:Willimas_NIPS01} for GMM (and RBF).  In this study, we will focus on the strategy based on GCWS hashing for linearizing the GMM  and NGMM kernels.

\subsection{GCWS: Generalized Consistent Weighted Sampling}

After we have transformed the data according to (\ref{eqn_transform}),   since the data are now nonnegative, we can apply the original ``consistent weighted sampling''~\cite{Report:Manasse_CWS10,Proc:Ioffe_ICDM10,Proc:Li_KDD15} to generate hashed data. ~\cite{Report:Li_GMM16} named this procedure  ``generalized consistent weighted sampling'' (GCWS), as summarized in Algorithm~\ref{alg_GCWS}.
\begin{algorithm}{
\textbf{Input:} Data vector $u_i$  ($i=1$ to $D$)

Generate vector $\tilde{u}$ in $2D$-dim by (\ref{eqn_transform}), then normalize $\tilde{u}$ so that $\sum_{i=1}^{2D}\tilde{u}_i = 1$.

\vspace{0.08in}

For $i$ from 1 to $2D$

\hspace{0.1in}$r_i\sim Gamma(2, 1)$, $c_i\sim Gamma(2, 1)$,  $\beta_i\sim Uniform(0, 1)$

\hspace{0.05in} $t_i\leftarrow \lfloor \frac{\log \tilde{u}_i }{r_i}+\beta_i\rfloor$, $z_i\leftarrow \exp(r_i(t_i - \beta_i))$,  $a_i\leftarrow c_i/(z_i \exp(r_i))$

End For

\textbf{Output:} $i^* \leftarrow arg\min_i \ a_i$,\hspace{0.3in}  $t^* \leftarrow t_{i^*}$
}\caption{Generalized consistent weighted sampling (GCWS) for hashing NGMM kernel}
\label{alg_GCWS}
\end{algorithm}

\noindent With $k$ samples, we  can  estimate $NGMM(u,v)$ according to the following collision probability: 
\begin{align}
&\mathbf{Pr}\left\{i^*_{\tilde{u},j} = i^*_{\tilde{v},j} \ \text{and} \ t^*_{\tilde{u},j} = t^*_{\tilde{v},j}\right\} = NGMM(u,v),
\end{align}

Note that  $i^*\in[1,\ 2D]$  and $t^*$ is unbounded.  Recently, \cite{Proc:Li_KDD15} made an interesting observation that for practical data, it is ok to completely discard $t^*$. The following approximation
\begin{align}\label{eqn_GCWS_Prob}
\mathbf{Pr}\left\{i^*_{\tilde{u},j} =  i^*_{\tilde{v},j}\right\} \approx NGMM({u},{v})
\end{align}
is  accurate in practical settings and makes the implementation  convenient via the idea of $b$-bit minwise hashing~\cite{Proc:Li_Konig_WWW10}.

For each  vector $u$, we obtain $k$ random samples $i^*_{\tilde{u},j}$, $j=1$ to $k$. We store only the lowest $b$ bits of $i^*$. We need to view those $k$ integers as locations (of the nonzeros) instead of numerical values. For example, when $b=2$, we should view $i^*$ as a  vector of length $2^b=4$. If $i^*=3$, then we code it as $[1\ 0\ 0\ 0]$; if $i^*=0$, we code it as $[0\ 0\ 0\ 1]$, etc. We  concatenate all $k$ such vectors into a binary vector of length $2^b\times k$, which contains exactly $k$ 1's.  After we have generated such new data vectors for all data points, we feed them to a linear classifier. We can, of course,  also use the new data for many other tasks including clustering, regression, and near neighbor search.

Note that for linear learning methods (especially online algorithms), the storage and computational cost is largely determined by the number of nonzeros in each data vector, i.e., the $k$ in our case. It is thus crucial not to use a too large $k$.

\subsection{An Experimental Study on ``0-bit'' GCWS}

\begin{figure}[h!]
\begin{center}
\mbox{
\includegraphics[width=2.3in]{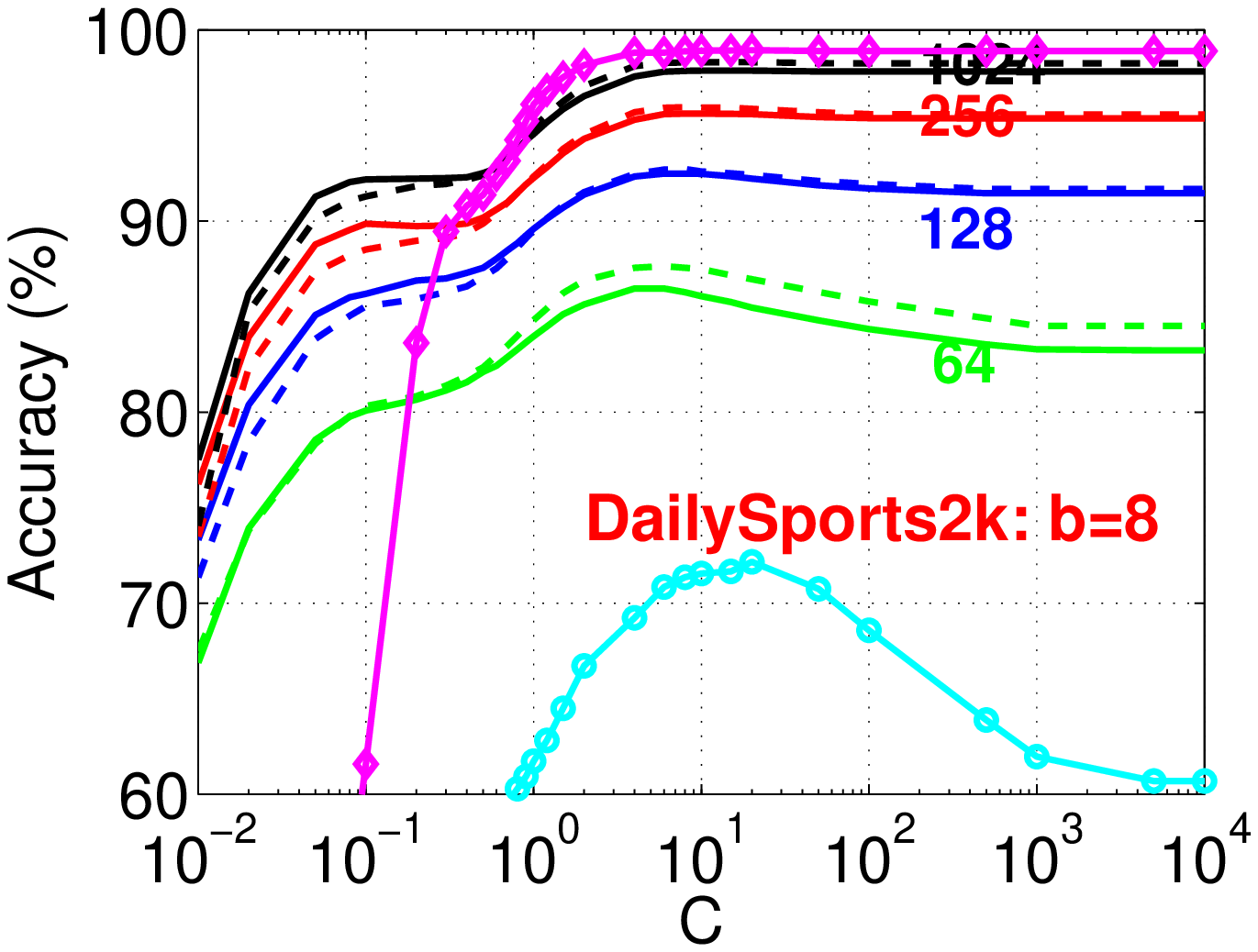}\hspace{-0.14in}
\includegraphics[width=2.3in]{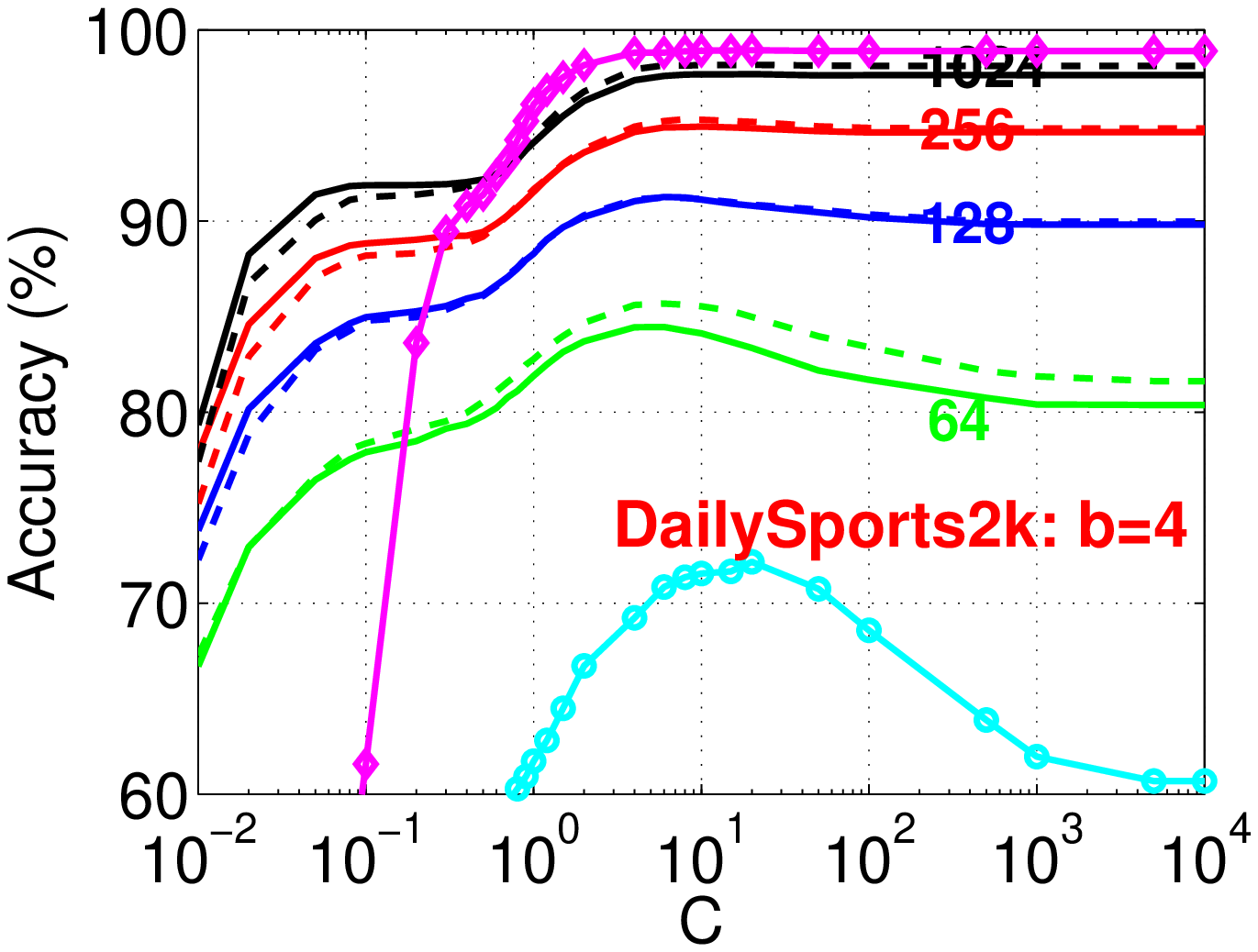}\hspace{-0.14in}
\includegraphics[width=2.3in]{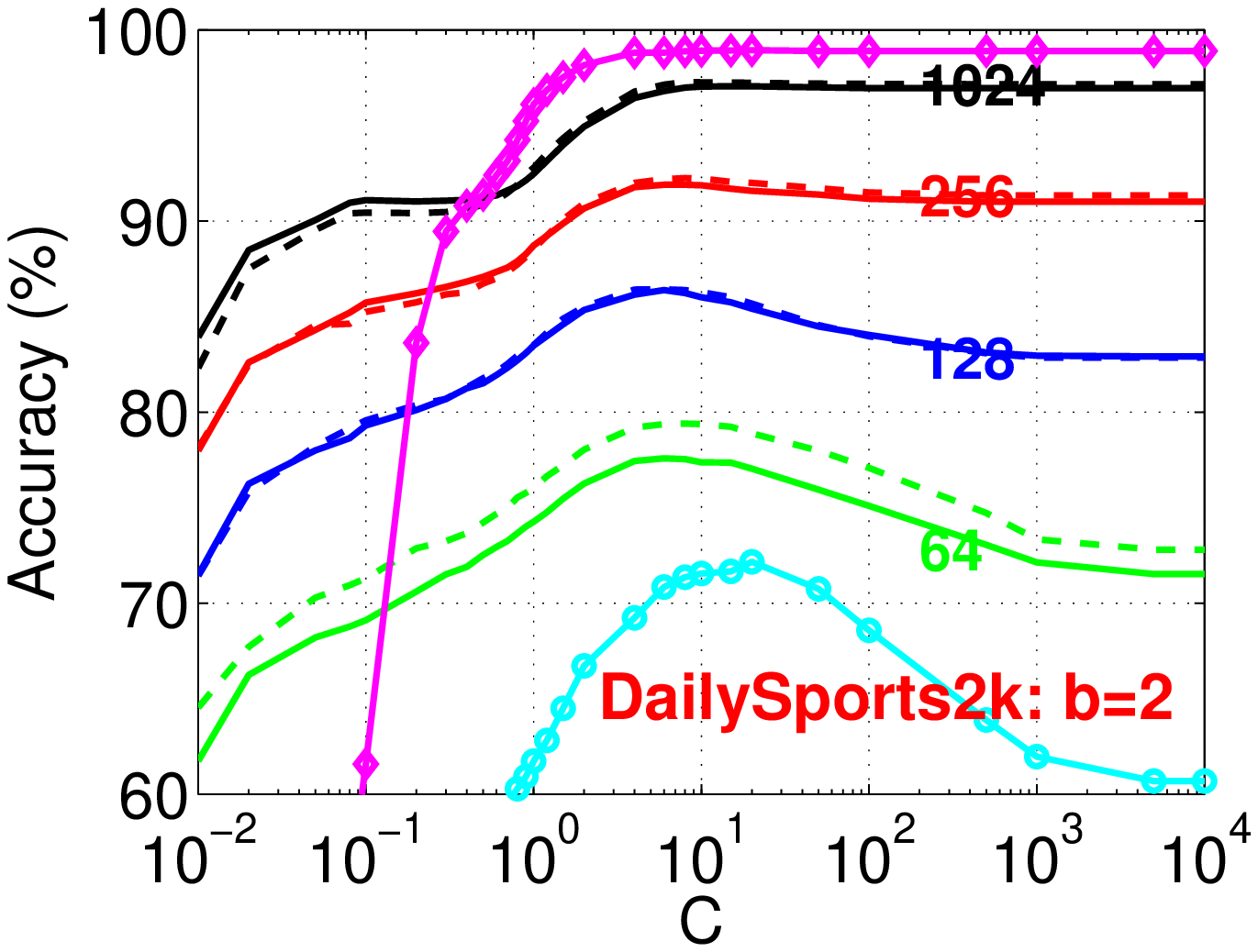}
}

\mbox{
\includegraphics[width=2.3in]{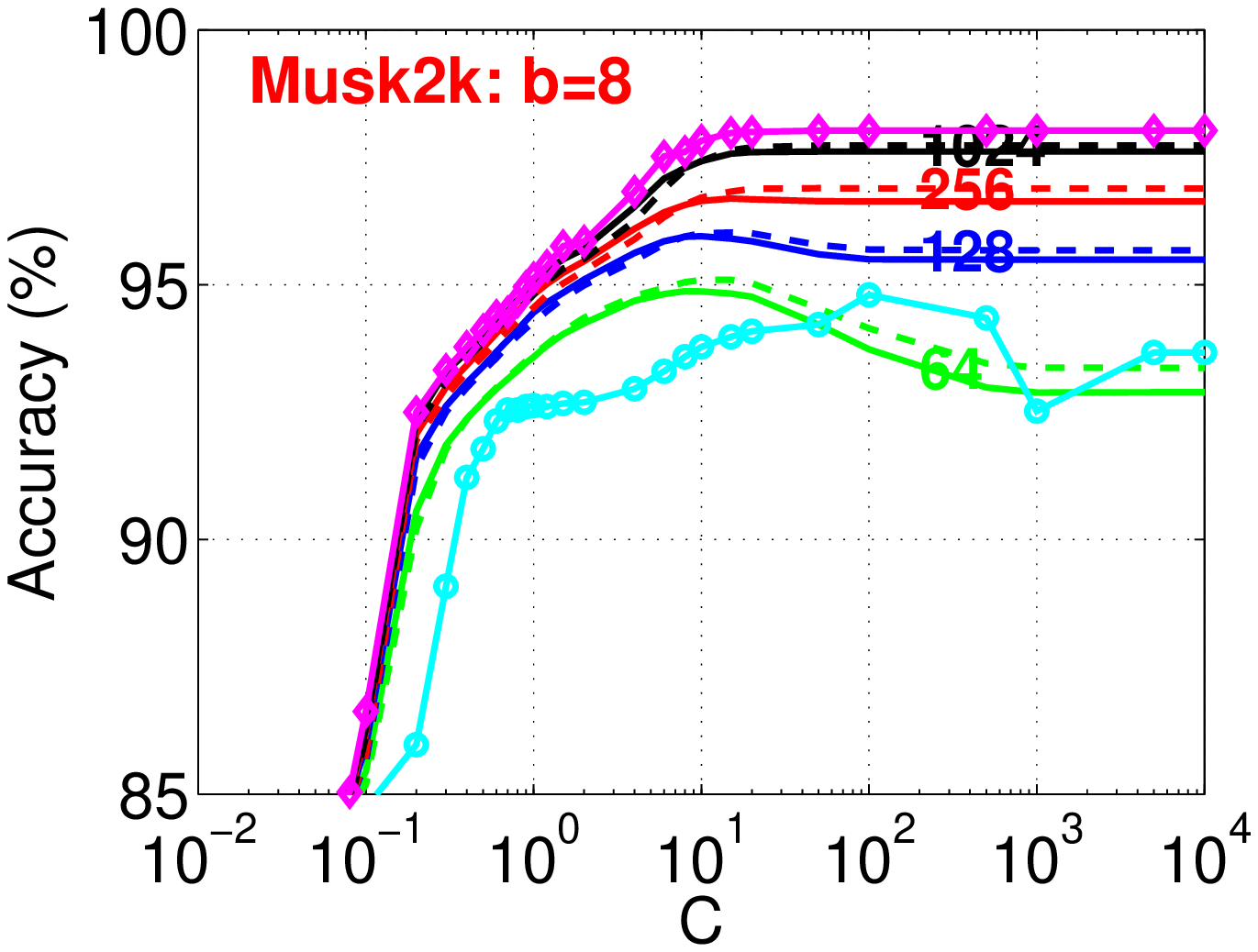}\hspace{-0.14in}
\includegraphics[width=2.3in]{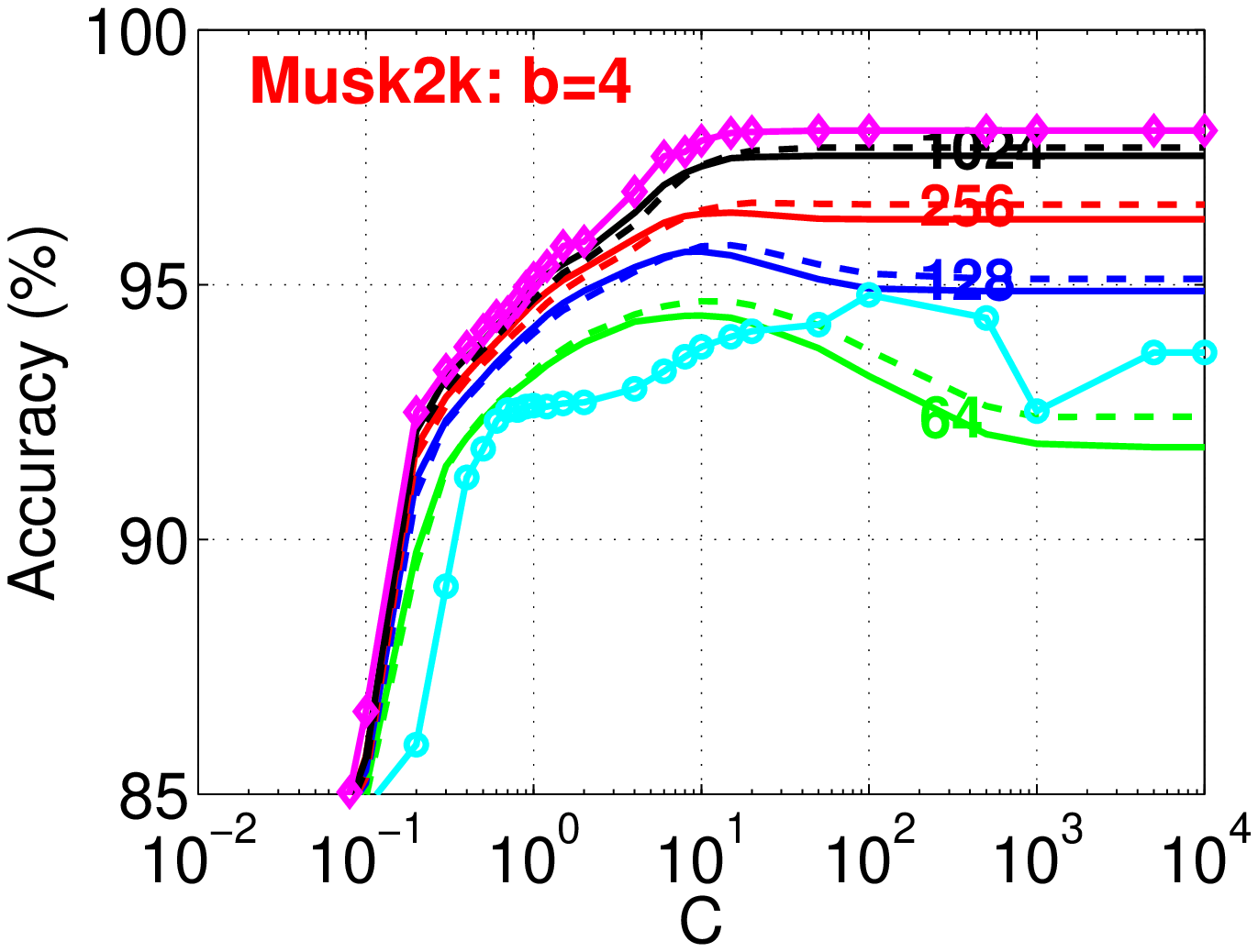}\hspace{-0.14in}
\includegraphics[width=2.3in]{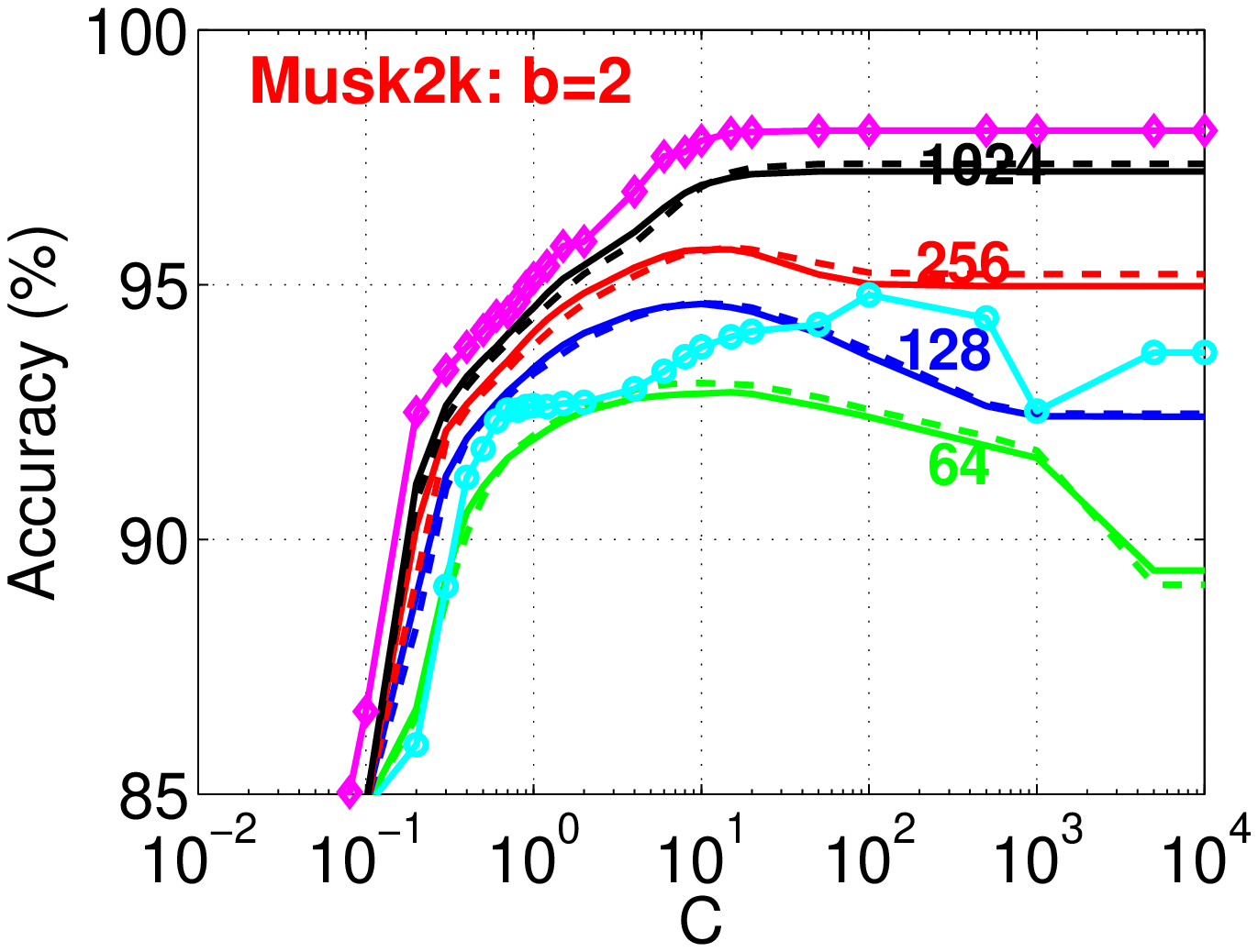}
}

\mbox{
\includegraphics[width=2.3in]{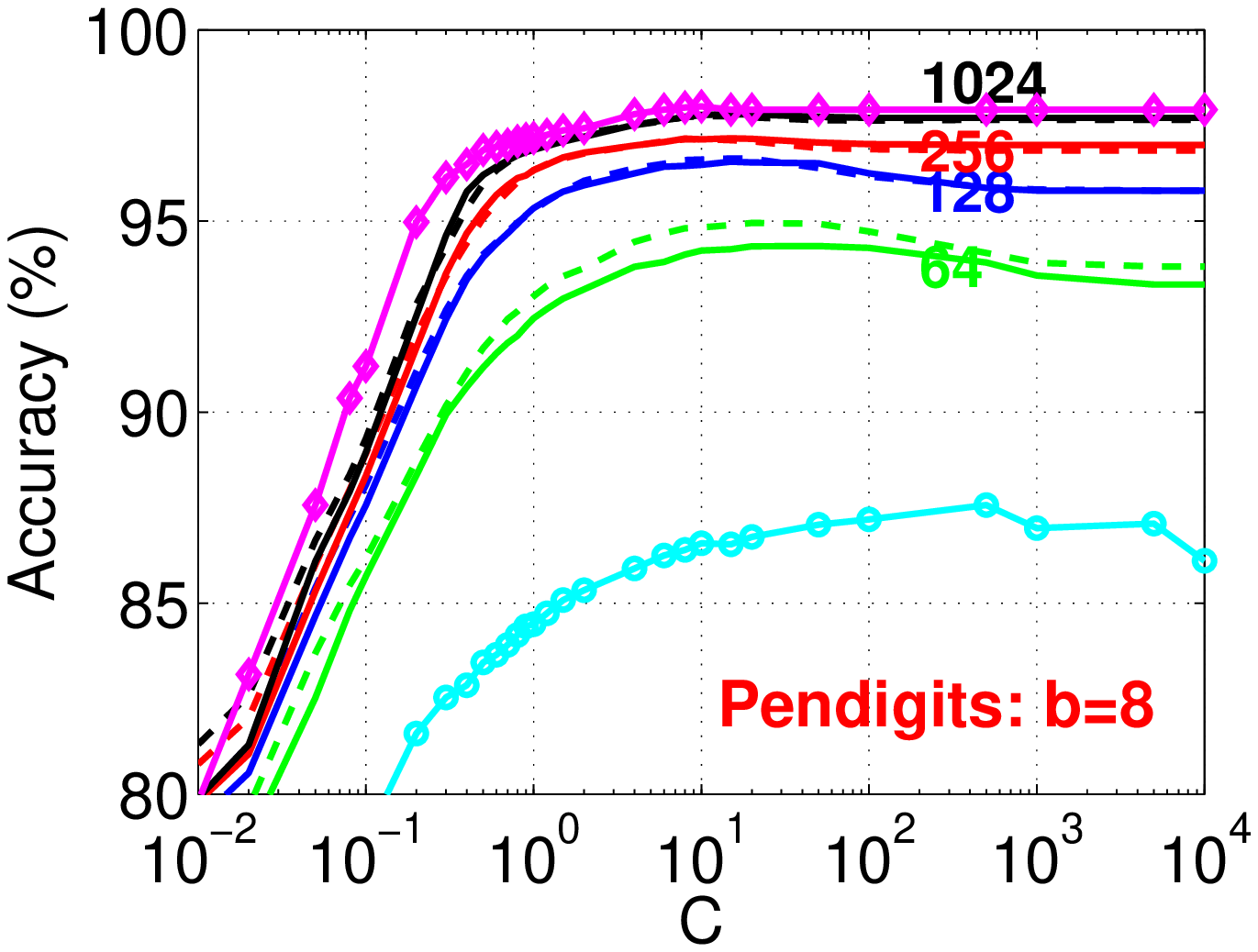}\hspace{-0.14in}
\includegraphics[width=2.3in]{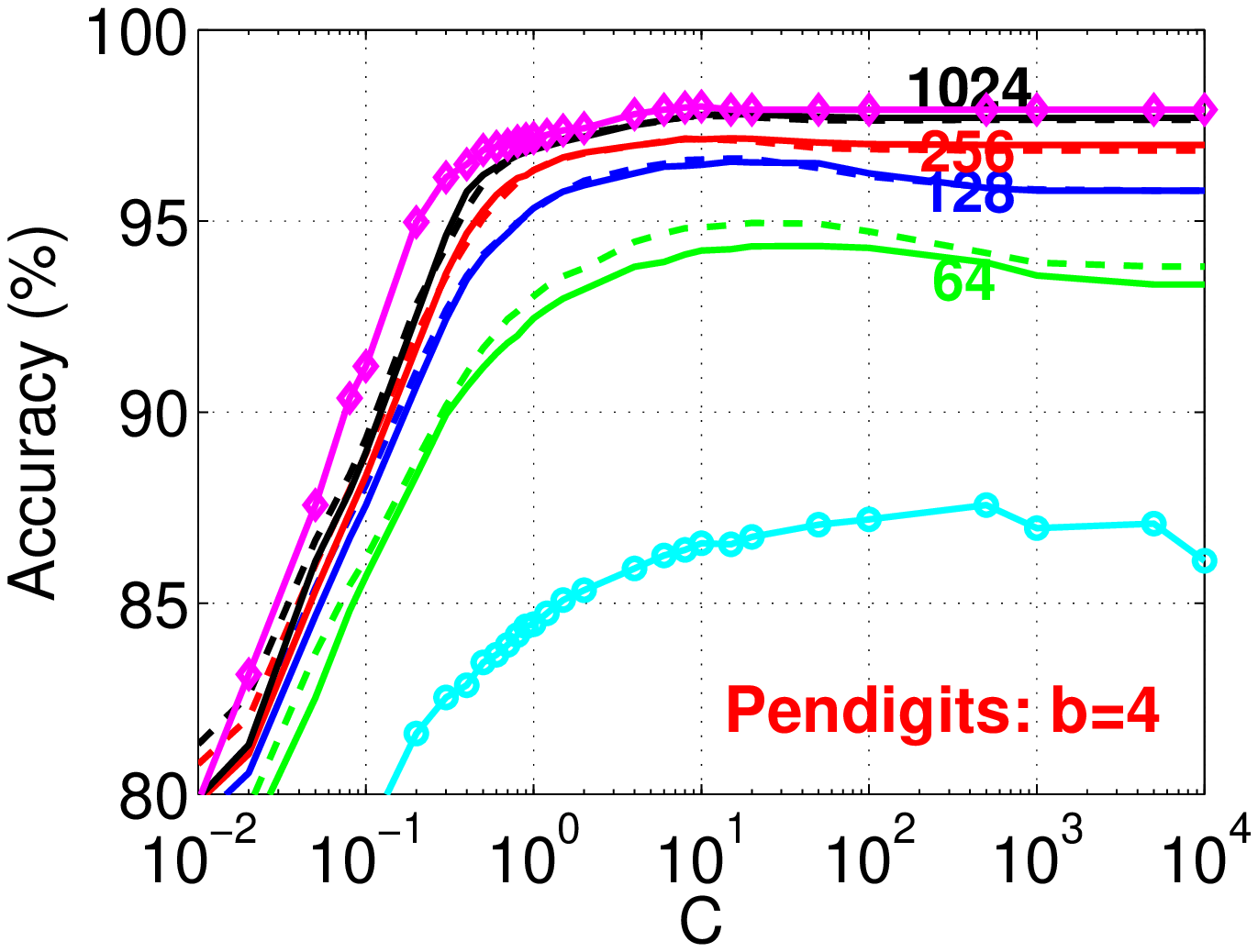}\hspace{-0.14in}
\includegraphics[width=2.3in]{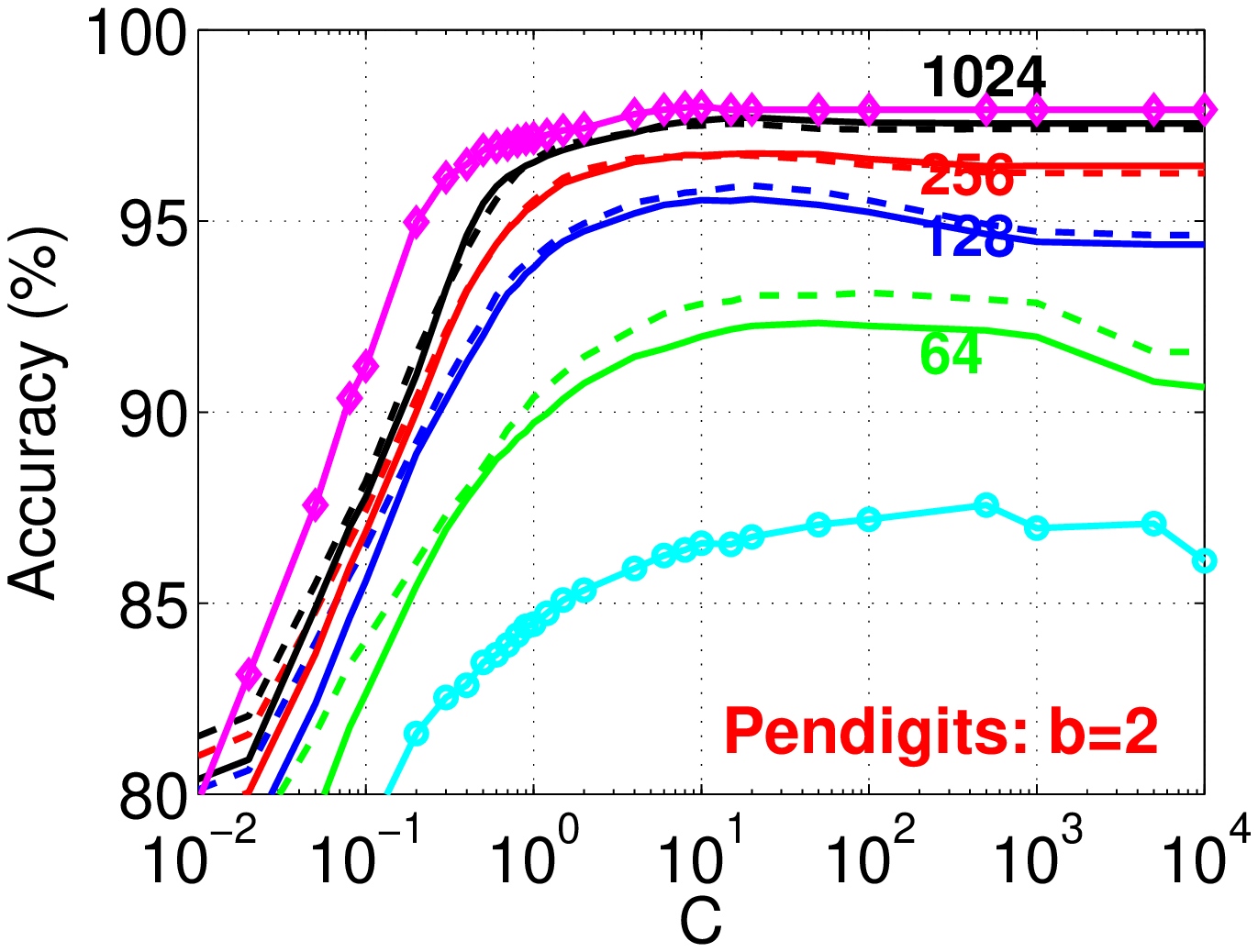}
}

\mbox{
\includegraphics[width=2.3in]{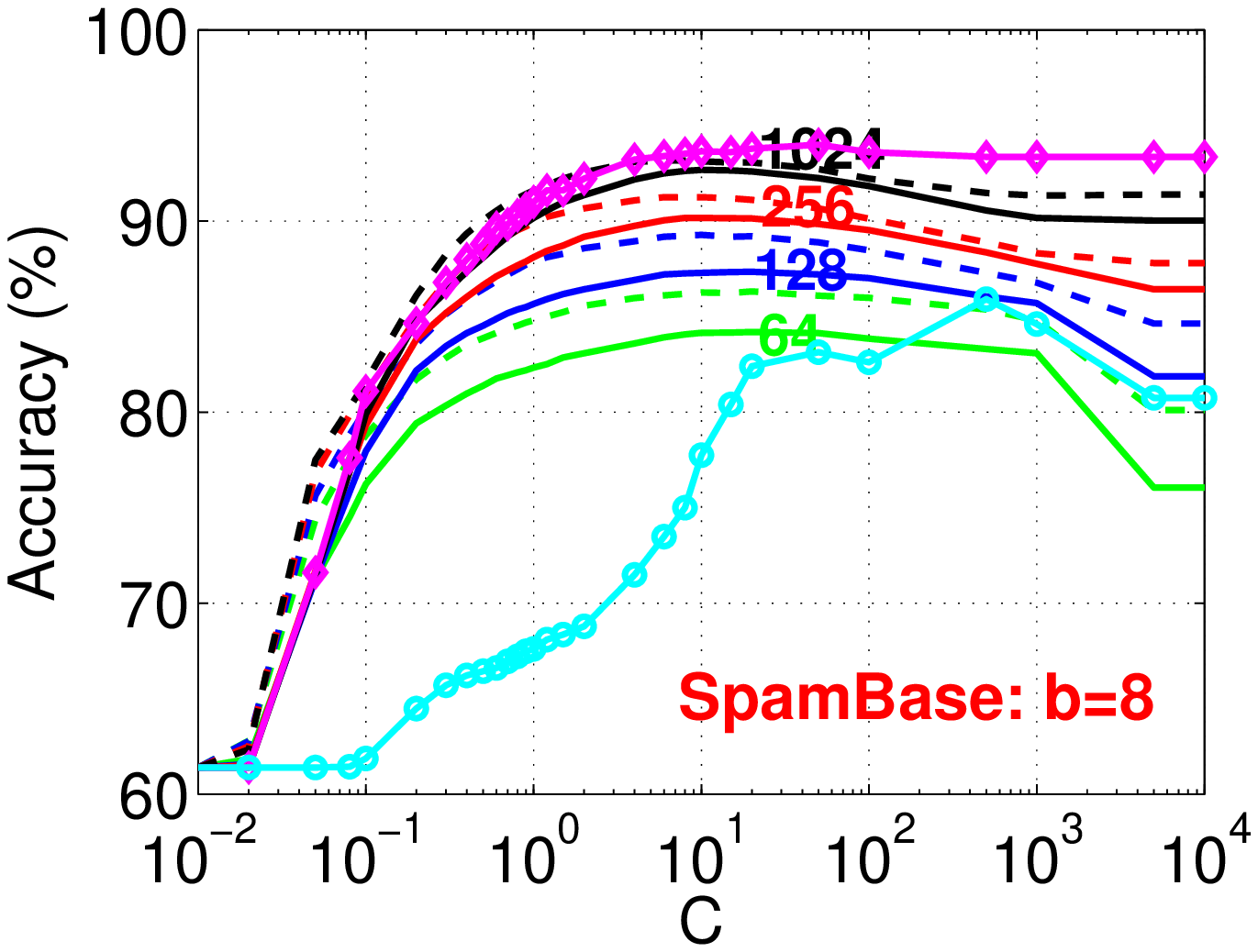}\hspace{-0.14in}
\includegraphics[width=2.3in]{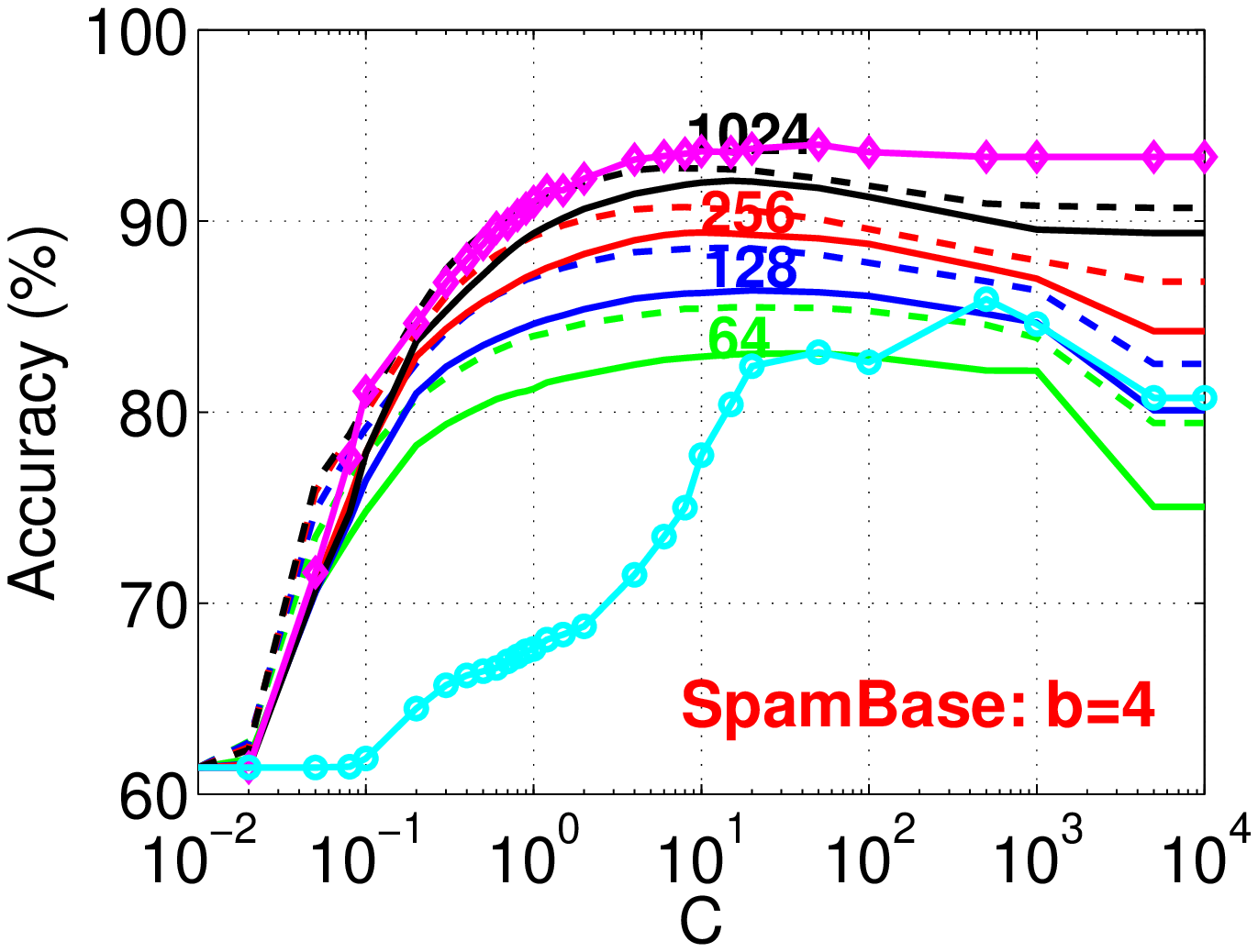}\hspace{-0.14in}
\includegraphics[width=2.3in]{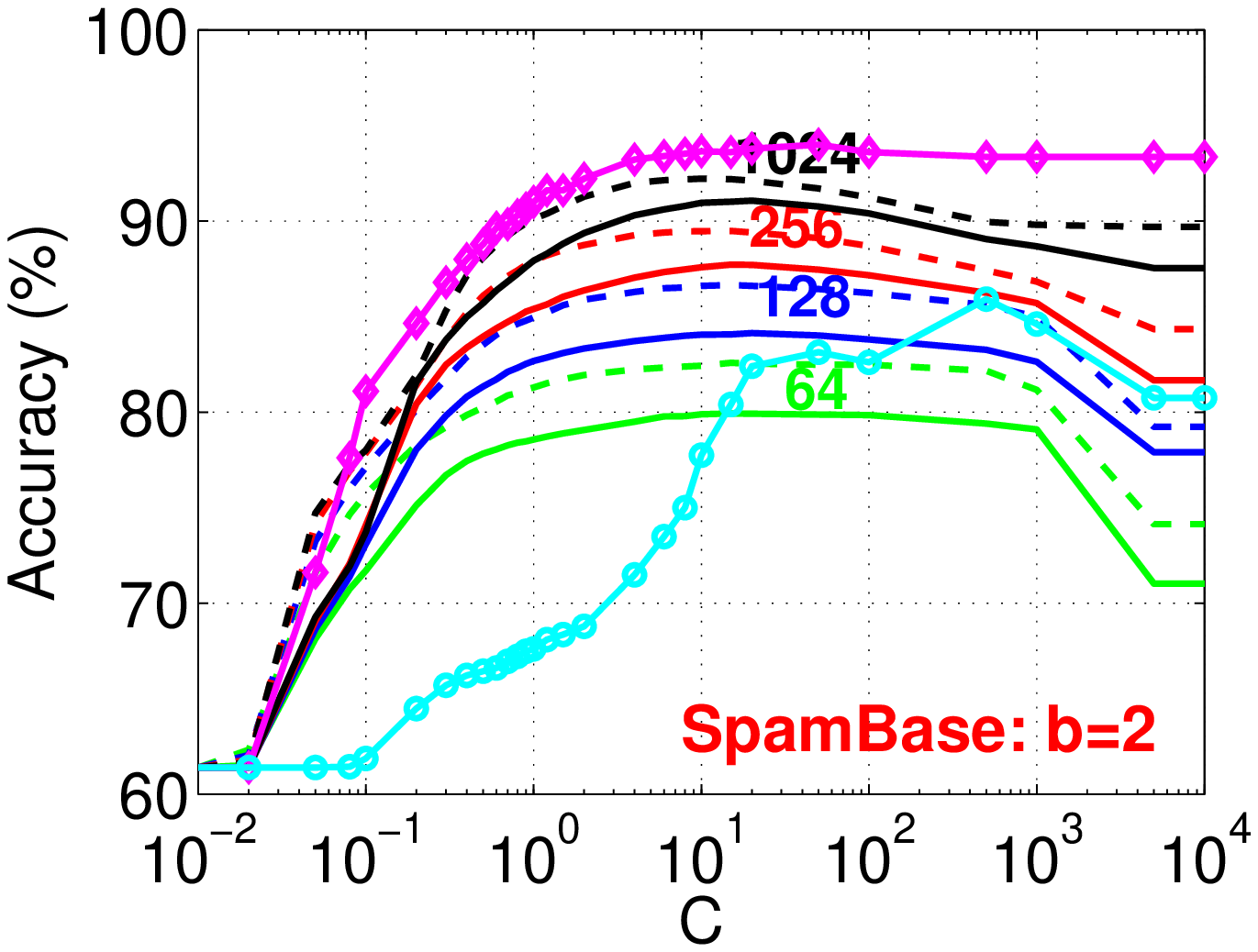}
}

\mbox{
\includegraphics[width=2.3in]{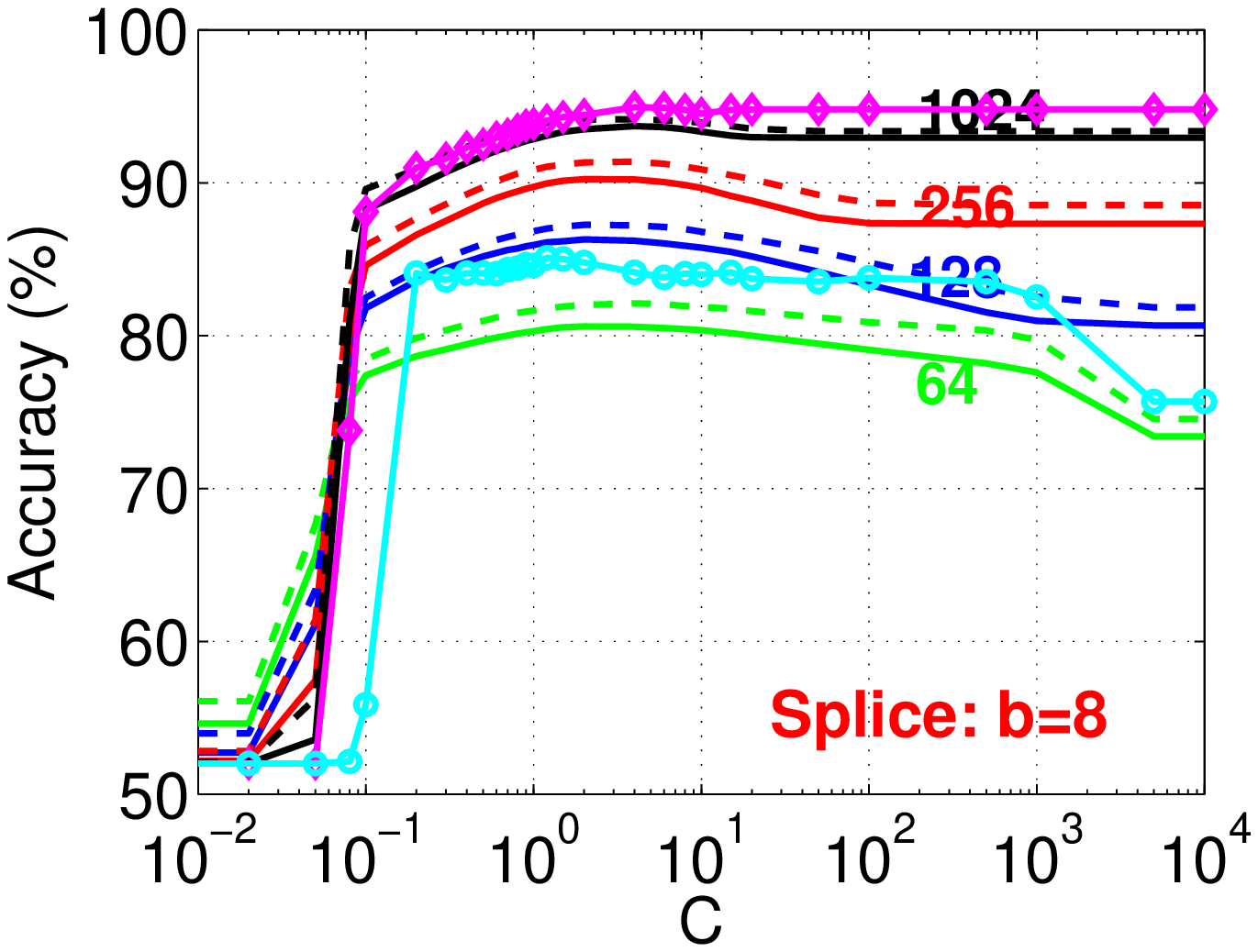}\hspace{-0.14in}
\includegraphics[width=2.3in]{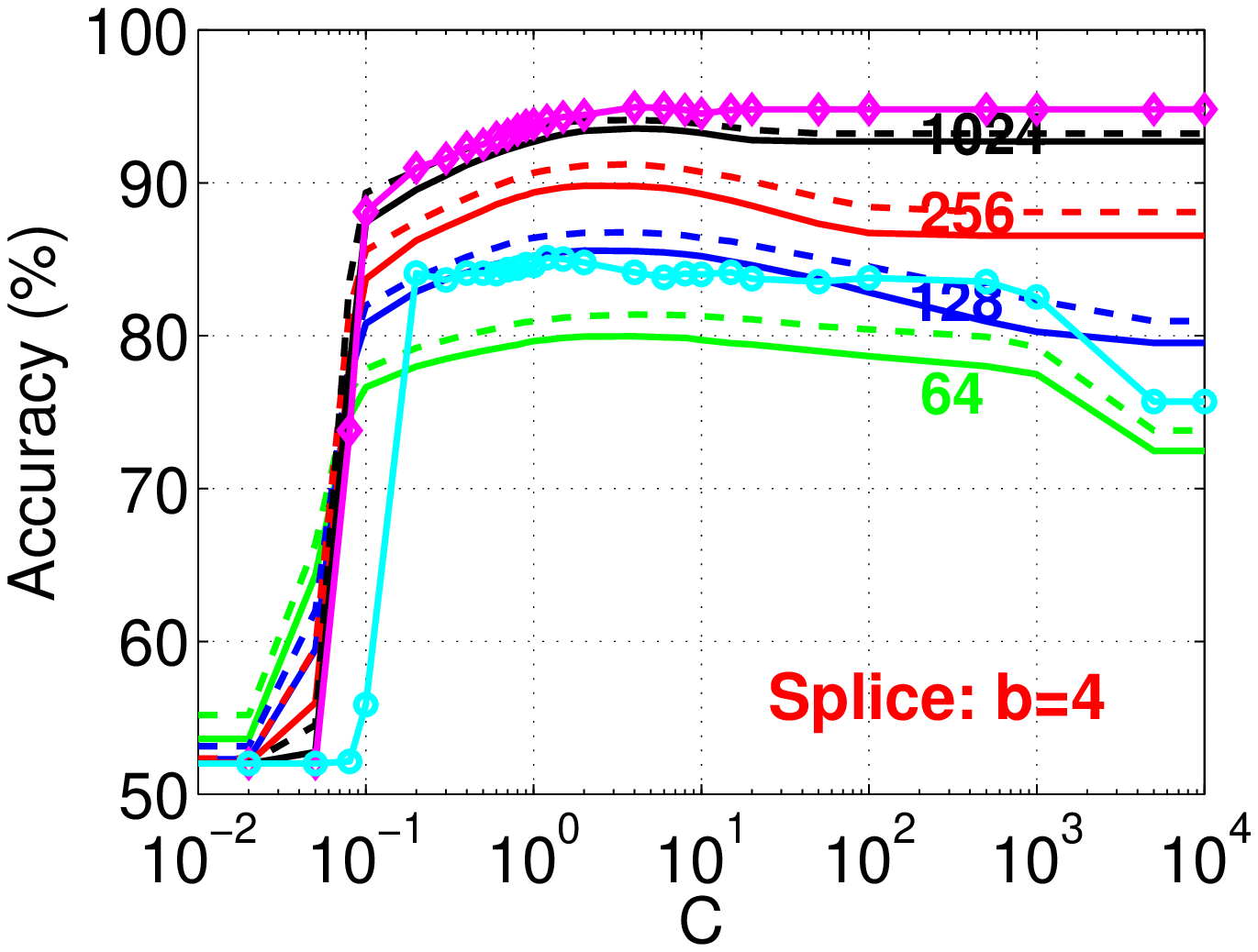}\hspace{-0.14in}
\includegraphics[width=2.3in]{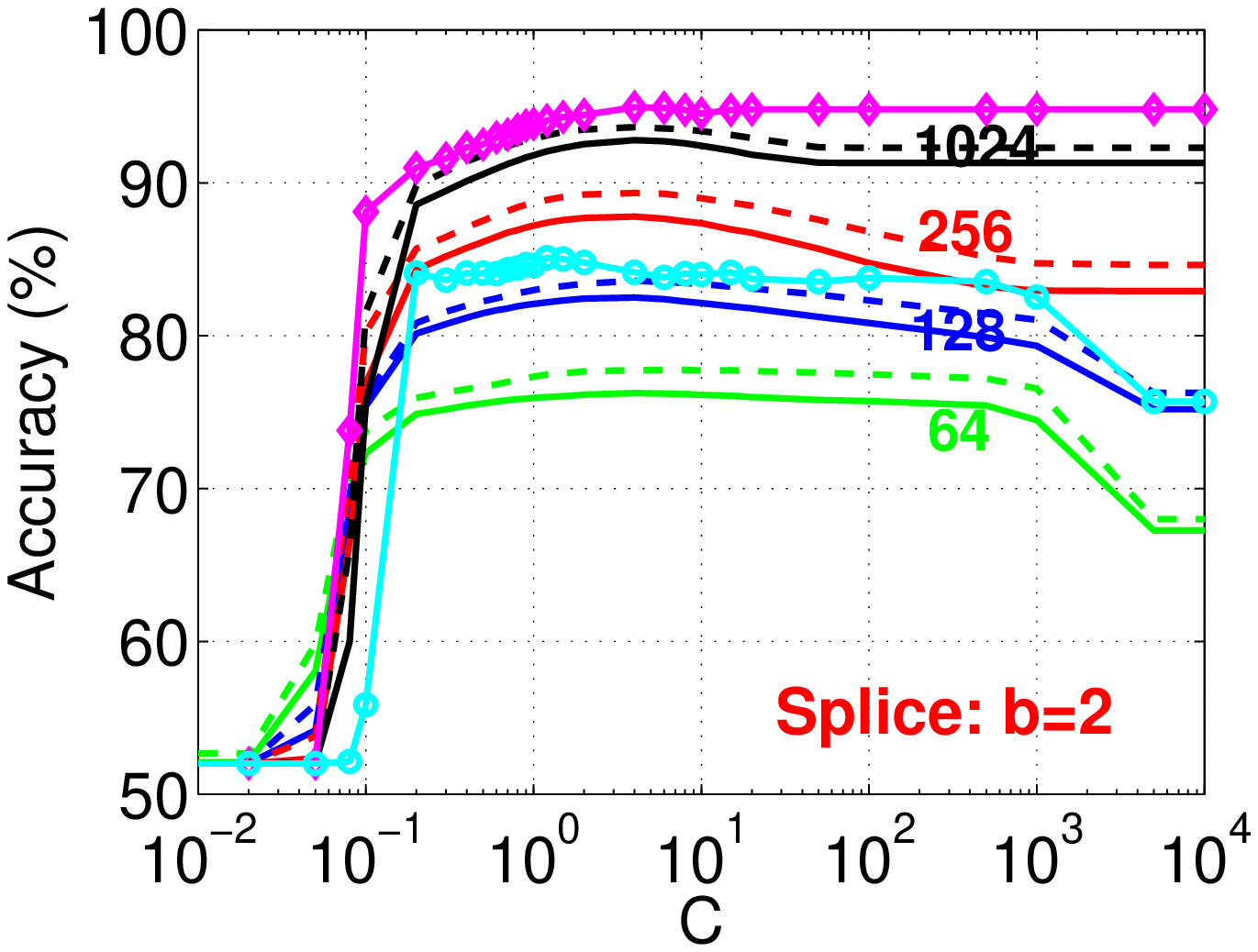}
}

\end{center}
\vspace{-0.3in}
\caption{Test classification accuracies for using (0-bit) GCWS to hash the NGMM kernel (solid curves) and GMM (dashed curves), for $k\in\{64,128,256,1204\}$ and $b\in\{8,4,2\}$. In each panel, the two solid and marked curves  report the  original results of the  NGMM kernel (upper curve) and the  linear kernel (bottom curve). }\label{fig_hash}

\end{figure}

Figure~\ref{fig_hash} reports the test classification accuracies on 0-bit GCWS for hashing the NGMM kernel (solid curves) and the GMM kernel (dashed curves). In each panel, we also report the original linear kernel and NGMM kernel results using two solid and marked curves, with the upper curve for the NGMM kernel and bottom curve for the linear kernel. We report results for $k\in\{64,128,256,1024\}$. We also need to choose $b$, the number of bits for encoding each hashed value $i^*$; we reports experimental results for $b\in\{8,4,2\}$.\\

The classification results confirm, just like the prior work~\cite{Report:Li_GMM16}, that the ``0-bit'' GCWS scheme performs well for hashing the NGMM kernel. Clearly, the accuracies are affected by the choice of $b$, but not too much especially when $k$ is not too small. In general, we recommend using a larger $b$ if the model size $2^b\times k$ is affordable. The   training cost is largely determined by $k$, not much by $b$. See~\cite{Report:Li_GMM16} for the training time  comparisons for different $b$ values. \\

Often times, practitioners are particularly interested in choosing $k$ large enough to exceed the accuracy of linear kernel. This is because in practice linear models are often used and a simple recipe which can be more accurate than linear models and does not increase much the computational cost would be highly desirable. We can see from Figure~\ref{fig_hash} that typically $k$ does not have to very large in order to outperform the original linear model.

\section{Conclusion}

We propose the ``generalized intersection (GInt)'' kernel and the related ``normalized generalized min-max (NGMM)'' kernel. The original (histogram) intersection kernel has been popular in (e.g.,) computer vision. Interestingly, the NGMM kernel can be viewed as an ``asymmetrically transformed'' version of the GInt kernel from the perspective of recently proposed ``asymmetric hashing''~\cite{Proc:ALSH_WWW15}. Our kernel SVM experiments on 40 UCI datasets illustrate that the NGMM kernel typically outperforms the GInt kernel. The recently proposed 0-bit GCWS  scheme performs well for approximating the NGMM kernel, as expected. For readers who are interested in Nystrom method (another  scheme for kernel linearization), please refer to an earlier technical report~\cite{Report:Li_GMM_Nys16}.

\vspace{0.1in}
\noindent In this study, we focus on reporting results for classification. Obviously, the techniques can be used for many other tasks including regression, clustering, and near neighbor search. One notable advantage of the (0-bit) GCWS is that the hashed values are of discrete nature and can be directly used for  building hash tables in the context of sublinear time near neighbor search. The Nystrom method dos not offer this benefit. Note that in the context of near neighbor search, the GCWS hashing  provides a scheme for searching near neighbors in terms of not only the NGMM kernel distance but also the GInt kernel distance, because NGMM is a monotonic transformation of GInt.

\vspace{0.1in}
\noindent While the results of this  (short) report appear to support more the GMM kernel than the GInt kernel, we hope these two  simple (tuning-free) kernels (NGMM and GInt) can provide practitioners more options to choose the appropriate method for their specific domain applications.

\newpage\clearpage

{
\bibliographystyle{abbrv}
\bibliography{../bib/mybibfile}

}

\end{document}